\documentclass[conference]{./support/ieeeconf}
\usepackage{times}
\usepackage{amsmath}
\usepackage[pdftex]{graphicx}
\usepackage{caption}
\usepackage{subfiles} 
\usepackage{subfigure}
\usepackage{graphicx}
\usepackage{amsmath,amssymb,amsopn,amstext,amsfonts}
\usepackage{cancel}
\usepackage[space]{cite}
\usepackage{pdfsync}
\usepackage{balance}
\usepackage{color}
\usepackage{algorithm}
\usepackage{algorithmic}
\usepackage{url}
\usepackage{booktabs}
\usepackage{threeparttable}
\usepackage[linkcolor=black,citecolor=black,urlcolor=black,colorlinks=true]{hyperref}
\usepackage{multirow}
\usepackage[outdir=./epstopdf/]{epstopdf}
\usepackage{graphicx}
\usepackage{array}
\usepackage{verbatim}
\usepackage{makecell}
\usepackage{subcaption}
\usepackage{stackengine}

\IEEEoverridecommandlockouts

\graphicspath{{./figure/}}

\DeclareGraphicsExtensions{.pdf,.png,.jpg,.eps,.PNG}
\title{\LARGE \bf Embodied Escaping: 
End-to-End Reinforcement Learning \\for Robot Navigation in Narrow Environment}
\author{Han Zheng$^{1\dagger}$, Jiale Zhang$^{1\dagger}$, Mingyang Jiang$^1$, Peiyuan Liu$^2$, Danni Liu$^2$, Tong Qin$^1$, and Ming Yang$^1$
    \thanks{
        $^{1}$Han Zheng, Jiale Zhang, Mingyang Jiang, Tong Qin, and Ming Yang are with Shanghai Jiao Tong University, Shanghai, China.
        {\tt\small \{zjlzjlqwq, jiamiya, qintong, MingYang\}@sjtu.edu.cn}.
    }
    \thanks{
        $^{2}$Peiyuan Liu and Danni Liu are with Cleanix Robotics Co., Ltd., China.
    }
    \thanks{
    {$\dagger$ means equal contribution.}
    }
    }

\begin{document}
\maketitle
\thispagestyle{empty}
\pagestyle{empty}

\begin{abstract}
Autonomous navigation is a fundamental task for robot vacuum cleaners in indoor environments.
Since their core function is to clean entire areas, robots inevitably encounter dead zones in cluttered and narrow scenarios. 
Existing planning methods often fail to escape due to complex environmental constraints, high-dimensional search spaces, and high difficulty maneuvers. 
To address these challenges, this paper proposes an embodied escaping model that leverages reinforcement learning-based policy with an efficient action mask for dead zone escaping.
To alleviate the issue of the sparse reward in training, we introduce a hybrid training policy that improves learning efficiency. 
In handling redundant and ineffective action options, we design a novel action representation to reshape the discrete action space with a uniform turning radius. 
Furthermore, we develop an action mask strategy to select valid action quickly, balancing precision and efficiency.
In real-world experiments, our robot is equipped with a Lidar, IMU, and two-wheel encoders. 
Extensive quantitative and qualitative experiments across varying difficulty levels demonstrate that our robot can consistently escape from challenging dead zones.
Moreover, our approach significantly outperforms compared path
planning and reinforcement learning methods in terms of success rate and collision avoidance. A video showcasing our methodology and real-world demonstrations is available at \url{https://youtu.be/kBaaYWGhNuE}.
\end{abstract}

\begin{figure}
\vspace{15pt}
	\centering
	\subfigure[The planning trajectory of the escaping task for differential robot]{
		\includegraphics[width=0.9\linewidth]{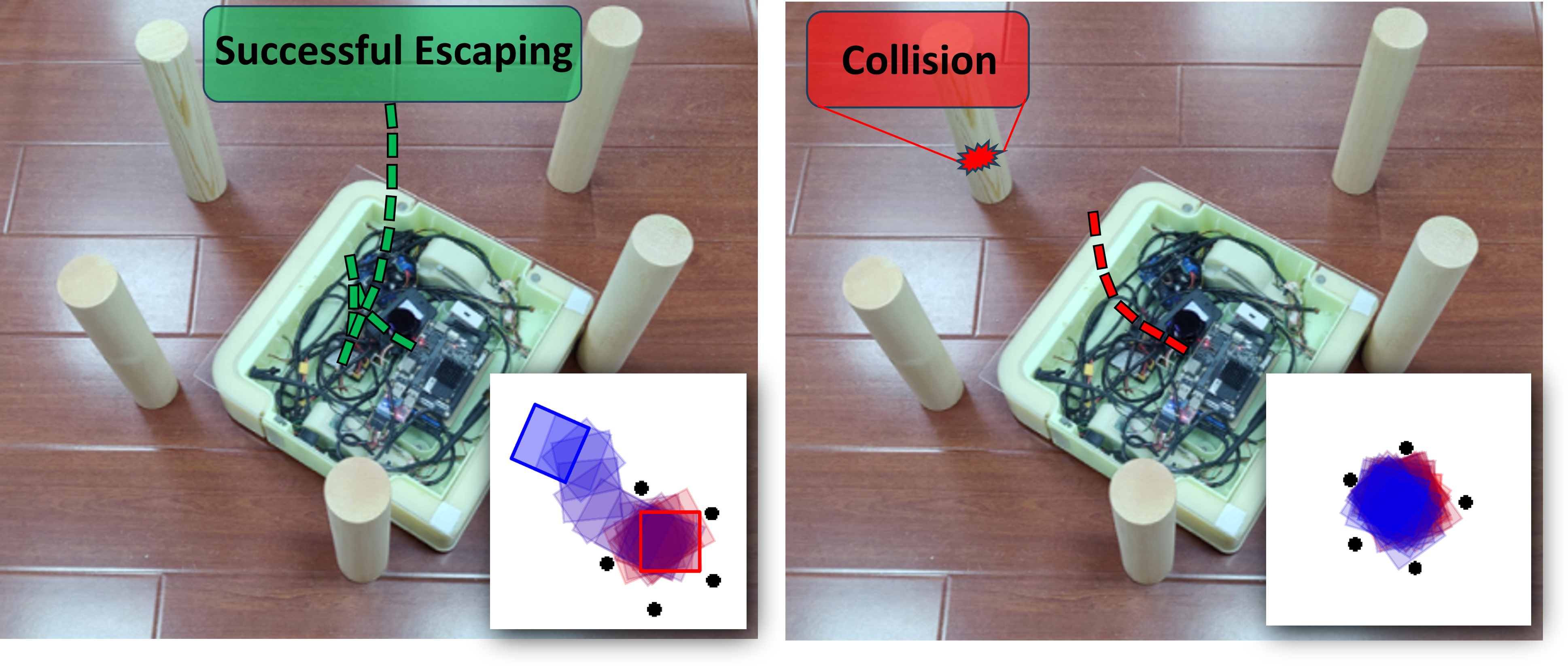} \label{fig:abs_a}
	}
	\subfigure[The basic idea of embodied escaping model]{
		\includegraphics[width=0.8\linewidth]{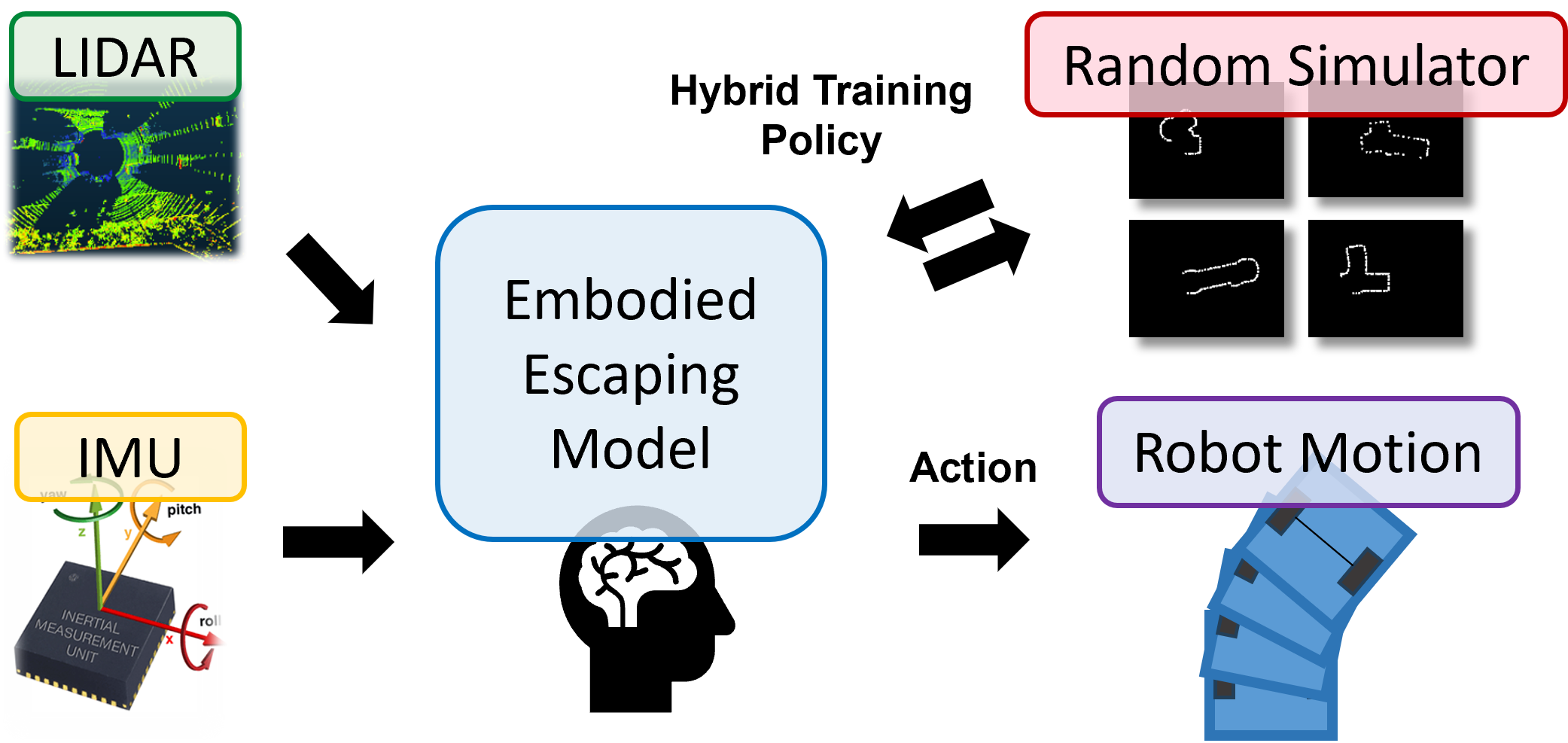} \label{fig:abs_b}
	}
	\caption{(a) shows an escaping process for robot
vacuum cleaners. (b) shows the basic idea of the embodied escaping model, which directly maps Lidar and IMU data to action. The end-to-end model is trained by the hybrid policy in a random simulator.}
\label{fig:abs}
\vspace{-10pt}
\end {figure}
 
\section{Introduction}
{P}{ath} planning in narrow environments has been extensively studied and applied to various real-world scenarios \cite{dobrevski2024dynamic,li2016time,pitkanen2021obstacle}, such as food delivery \cite{antony2018food}, disaster rescue \cite{murphy2004human}, and home cleaning \cite{jokic2022semantic}. 
For robotic vacuum cleaners, they operate in cluttered and confined indoor environments, which are prone to getting stuck in dead zones.
Many path planning algorithms focus on identifying and avoiding these zones \cite{zhang2010real,tan2021comprehensive} to prevent mobile robots from becoming stuck. 

However, the primary task of a robotic vacuum cleaner is to cover every corner of a room for cleaning, making it unavoidable for the robot to enter dead zones. 
Once encountered, continuous adjustments to the robot's pose during cleaning make it easy to enter but challenging to exit. 
When the path planning module repeatedly fails to generate a feasible path, the system concludes that the robot is trapped and requires an escape mechanism. 
However, current path planning and reinforcement learning methods often fail to extricate the robot under such conditions for several reasons \cite{yasuda2023safe,jian2021global}.
One challenge arises from the highly constrained environments. 
Household floors are typically non-convex, narrow, and cluttered with obstacles. 
These conditions make it difficult to maintain sufficient free space with geometric approximations, significantly increasing the risk of collisions \cite{shen2023multi}.
Furthermore, the accuracy of environmental modeling becomes a critical issue. 
Robot navigation usually relies on precise mapping, meaning that even slight errors in the environmental model can have a profound impact on the success of path planning \cite{zheng2023fast}.
Additionally, the nonlinear and non-holonomic robot dynamics cannot be neglected. 
As a result, in continuous high-dimensional spaces, the feasible trajectories for successful escape are extremely sparse, making the search for possible solutions particularly difficult \cite{yan2023learning}. 
Besides, sharp attitude changing poses further difficulty. 
Escaping from confined spaces often requires complex maneuvers, including frequent changes in direction and saturated turning rates, leading to trajectories filled with sharp turns and abrupt turning points \cite{shen2023multi}. 
Compounding these challenges, the task becomes extremely difficult. 

To address these problems, we have developed an end-to-end embodied escape model, allowing the robotic vacuum cleaner to navigate effectively out of the traps. 
The contributions of this paper are summarized as follows:
\begin{itemize}
	\item We designed an end-to-end and map-free escape architecture based on reinforcement learning. To alleviate sparse rewards associated with long-distance navigation, we employed a hybrid training policy, using A* to accelerate convergence near target points.
	\item We innovatively proposed an efficient action representation by proportionally scaling linear and angular velocities, reshaping a discrete action space with a uniform turning radius. A fast action mask is developed to model effective action distributions.
	\item 
 We deployed the proposed algorithm into real-world scenarios. 
 Challenging real-world experiments successfully verify the generalization ability.
 Comparative experiments demonstrate that our method significantly outperforms others. 
\end{itemize}   

\section{literature review}

\subsection{Rule-based Path Planning}
Rule-based path planning methods can be broadly categorized into three types: search-based, sampling-based, and optimization-based approaches \cite{li2021optimization}.

Search-based methods operate by discretizing the state or control space into finite elements and converting them into a connected graph of nodes and edges. In addition to the classical A* algorithm \cite{hart1968formal}, numerous variants based on A* have been proposed \cite{li2023app}. However, the above methods ignore the kinetic constraints. Among control space search methods, the Dynamic Window Approach (DWA) is particularly notable \cite{fox1997dynamic}. 
When control inputs contain errors, Yasuda et al. approximate the final state of each candidate input as a one-dimensional distribution, generating deterministic sample paths to reduce errors and enhance the safe control of a differential drive robot\cite{yasuda2023safe}.

Sampling-based methods are best represented by the Rapidly-exploring Random Tree (RRT) \cite{noreen2016optimal}. Noreen et al. designed an RRT-based global planner (RTR)  that generates paths consisting of straight-line motions and in-place turns, well-suited for differential-drive robots \cite{nagy2015path}. To further improve efficiency in dense environments and adaptability to narrow passages, an online path planning algorithm for Ackermann-steered vehicles based on RRT has been developed \cite{peng2021towards}. However, RRT still faces challenges in narrow environments when dealing with kinematic constraints.

Optimization-based methods model the planning problem using continuous variables, providing a more precise and unified task description. Pioneering work in autonomous parking introduced a novel convex optimization approach, leveraging strong duality to transform non-smooth collision avoidance constraints into smooth and differentiable ones \cite{zhang2020optimization}. Due to the differences between Ackermann and differential steering kinematics, Peng et al. proposed Kinetic Energy Difference (KED) as a measure of collision proximity \cite{pitkanen2021obstacle}. 
However, narrow spaces may significantly increase the search time for generating initial guesses, which is critical for optimization.

\subsection{Reinforcement Learning for Path Planning}
Compared to traditional rule-based path planning, reinforcement learning (RL) paradigms \cite{arulkumaran2017deep} are renowned for their superior adaptability, making them a prominent research direction in the field of robotic navigation. 
Nonetheless, RL methods still exposed the limitations of long-distance navigation in complex environments. 

A hierarchical reinforcement learning method was proposed, combining semantic knowledge with instance-based navigation points to tackle long-horizon tasks efficiently \cite{schmalstieg2023learning}. 
To address the shortsighted issue, methods using sub-goal generation \cite{zhu2022hierarchical,kastner2021connecting} have also been widely studied to reduce the state space and mitigate sparse rewards, significantly improving the sampling efficiency and generalization capacity. 
To adapt to multiple highly dynamic and cluttered environments, spatial and temporal re-planning mechanisms were added to the planner \cite{kastner2021arena}. 
In contrast to methods that rely on sub-goal suggestions, a unified diffusion policy was introduced to handle both goal-directed navigation and goal-agnostic exploration \cite{sridhar2024nomad}. 
This unified strategy demonstrated superior overall performance when navigating to visually indicated targets in constrained spaces.

Curriculum learning has been shown to be highly effective in accelerating convergence and improving performance across diverse tasks. 
One study developed a curriculum with distance and velocity functions that adapt to environmental difficulty, addressing the sparsity of positive learning samples \cite{wang2023curriculum}. 
However, the above RL-based methods currently suffer from low exploration efficiency and high collision risk in confined environments, making them unsuitable for tasks like robot escape.

\begin{figure*}[t]
	\centering
	\includegraphics[width=0.98\textwidth]{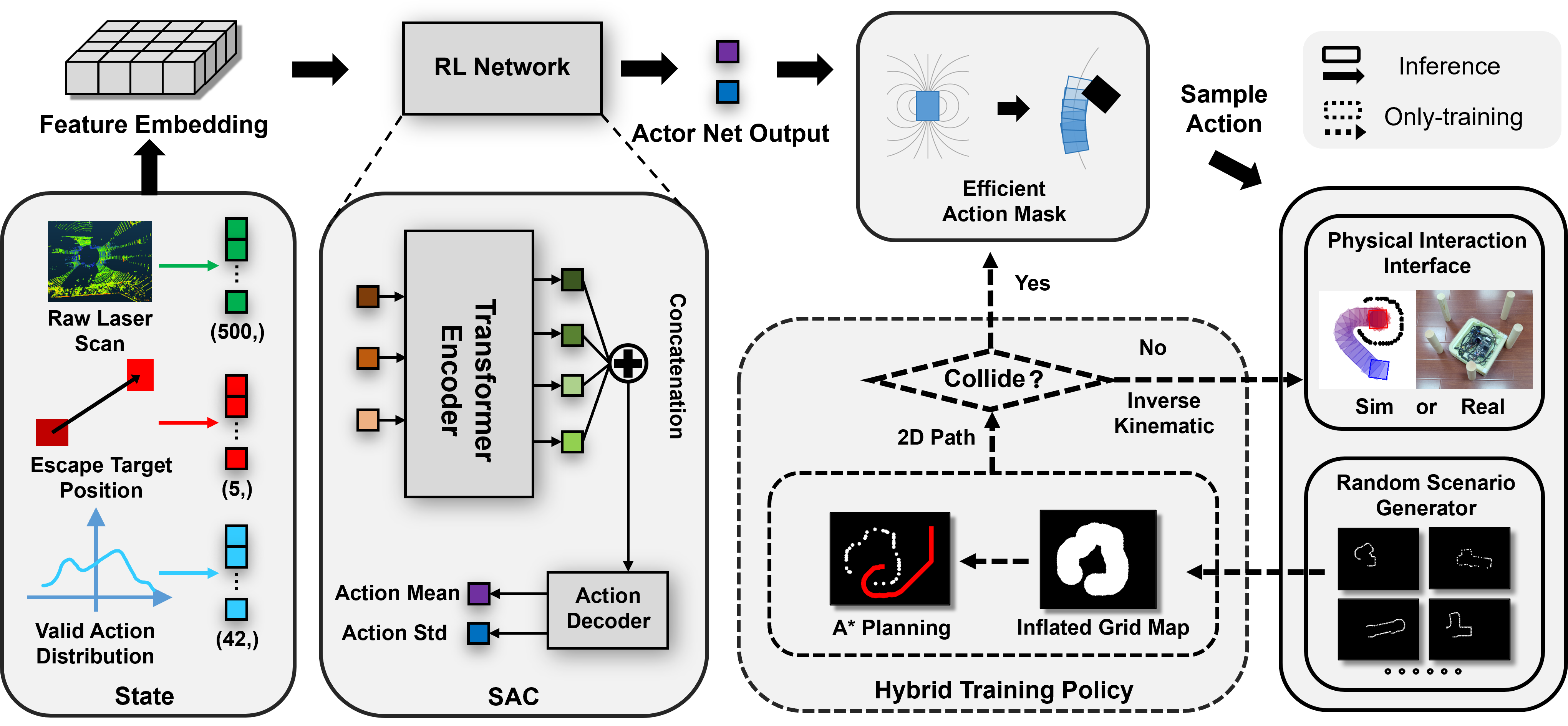}
	\caption{
The structure of the proposed embodied escaping model. 
The observation of the state space is elaborated in Sec. \ref{sec: problem}. Then the RL network is based on a SAC model with a transformer encoder. The hybrid training policy is illustrated in Sec. \ref{sec: hybrid}, which is trained in a 2D simulator. An efficient action mask is introduced in Sec. \ref{sec: action_mask}.}
\label{fig:framework}
\end{figure*}

\section{methodology}
We designed an end-to-end based framework, which directly maps raw sensor data inputs to control signals. 
The overall system is built upon the Soft Actor-Critic (SAC) \cite{haarnoja2018soft} and consists of two key components: 
(i) a reinforcement learning training process with a hybrid strategy, which improves success rates and facilitates precise arrival at the target destination; 
And (ii) an efficient action space representation enables the implementation of the action mask, significantly enhancing exploring safety and training efficiency.
The structure of the proposed embodied escape model is illustrated in Fig. \ref{fig:framework}. 
In the following sections, we will first formulate the problem and then explain each component.

\subsection{Problem Formulation}
\label{sec: problem}
For the escape task in narrow environments, the robot needs to achieve both obstacle avoidance and successful escape. 
We formalize this task as a Markov Decision Process (MDP) defined by the tuple $\langle S, A, T, R, \gamma \rangle$, where $S$ is the state space, $A$ is the action space, $T: S \times A \times S \rightarrow \mathbb{R}$ is the state transition model, $R: S \times A \rightarrow \mathbb{R}$ is the reward function, $\gamma \in [0,1)$ is the discount factor. 
At each time step $t$, the agent occupies a state $s_t \in S$ and selects an action $a_t \in A$ based on its policy $\pi(a_t | s_t)$. 
The agent then receives a reward $R(s_t, a_t)$ and transite to a new state $s_{t+1}$ according to the conditional transition probability $P(s_{t+1} | s_t, a_t)$. 
The goal of the MDP is to find a policy $\pi$ that maximizes the expected cumulative reward:
\begin{equation}
\pi^* = \arg \max_{\pi} \mathbb{E}_{s_0, a_0, \dots} \left[ \sum_{t=0}^{\infty} \gamma^t R(s_t, a_t) \right]
\end{equation}
\subsubsection{State Representation}
In the escape task, the observation of the state space comprises three key components, each is a flattened vector that is processed by an MLP and then concatenated to form the feature embedding.

\begin{itemize}
\item \textbf{Raw Laser Scan:} The input consists of a single-line 2D laser scan, which captures a dense ring of 500 points per full 360-degree sweep.

\item \textbf{Escape Target Position:} The escape target position is defined as a five-element tuple represented by
$(d, \cos(\theta_t), \sin(\theta_t), \cos(\phi_t), \sin(\phi_t))$. Here, $d$ represents the distance to the target in the ego robot's coordinate system, $\theta_t$ is the orientation angle, and $\phi_t$ is the heading angle.

\item \textbf{Valid Action Distribution:} The valid action distribution identifies and filters out unsafe actions using an action mask, which calculates 42 discrete actions. This process will be detailed in Sec. \ref{sec:reshape}.
\end{itemize}
\subsubsection{Action Space and State Transition}
The action space of a differential-drive robot is two-dimensional, consisting of angular velocity and linear velocity. And the set of all possible actions at each time step forms a joint action space \( A \), which is defined as:
\begin{equation}
A = \left\{ (\omega, v) \; | \; \omega \in [-\omega_{\max}, \omega_{\max}], v \in [-v_{\max}, v_{\max}] \right\}
\end{equation}
If the orientation at time \(t_i\) is \(\theta(t_i)\), the state transition can be derived during the given interval\((t_{i},t_{i+1})\):
\begin{equation}
\begin{cases}
    \theta(t_{i+1}) = \theta(t_i) + \omega_i  (t_{i+1}-t_{i}) \\
    x(t_{i+1}) = x(t_i) + \frac{v_i}{\omega_i} \left( \sin(\theta(t_{i+1})) - \sin(\theta(t_i)) \right) \\
    y(t_{i+1}) = y(t_i) - \frac{v_i}{\omega_i} \left( \cos(\theta(t_{i+1})) - \cos(\theta(t_i)) \right)
\end{cases}
\end{equation}
where radius \( r = \frac{v_i}{\omega_i} \) assuming \(\omega_i \neq 0\). If \(\omega_i = 0\), the robot will move along a straight line in the direction \(\theta(t_i)\).
At each control interval, the 2D search space is reduced to discrete lines space in Sec.  \ref{sec:reshape}.
\subsubsection{RL Network}
In our RL network, we build upon the SAC algorithm while integrating a custom architecture. SAC is a maximum-entropy reinforcement learning method that optimizes a stochastic policy by maximizing both the expected return and the policy entropy. Specifically, the objective is to maximize \(V^\pi(s)\), which is defined as:
\begin{equation}
\begin{aligned}
V^\pi(s) 
&= \mathbb{E}_{a \sim \pi(\cdot \mid s)}\bigl[Q^\pi(s, a) + \alpha\,H\bigl(\pi(\cdot \mid s)\bigr)\bigr] \\
&= \mathbb{E}_{a \sim \pi(\cdot \mid s)}\bigl[Q^\pi(s, a) - \alpha \,\log \pi(a \mid s)\bigr]
\end{aligned}
\end{equation}
where \(V^\pi(s)\) denotes the soft state value function at state \(s\) under policy \(\pi\); \(Q^\pi(s, a)\) is the soft state-action value function; \(\pi(a \mid s)\) represents the probability of selecting action \(a\) in state \(s\); \(\alpha\) is the temperature parameter balancing reward and exploration; and \(H\bigl(\pi(\cdot \mid s)\bigr)\) is the entropy of the policy at \(s\).

Our architecture enhances the standard SAC model by incorporating a transformer-based encoder to fuse feature embeddings, followed by an MLP that decodes these representations into actor net output as final action distribution.

\begin{figure}[t]
	\centering
	\subfigure[The example of A* planning result.]{
		\includegraphics[width=.8\linewidth]{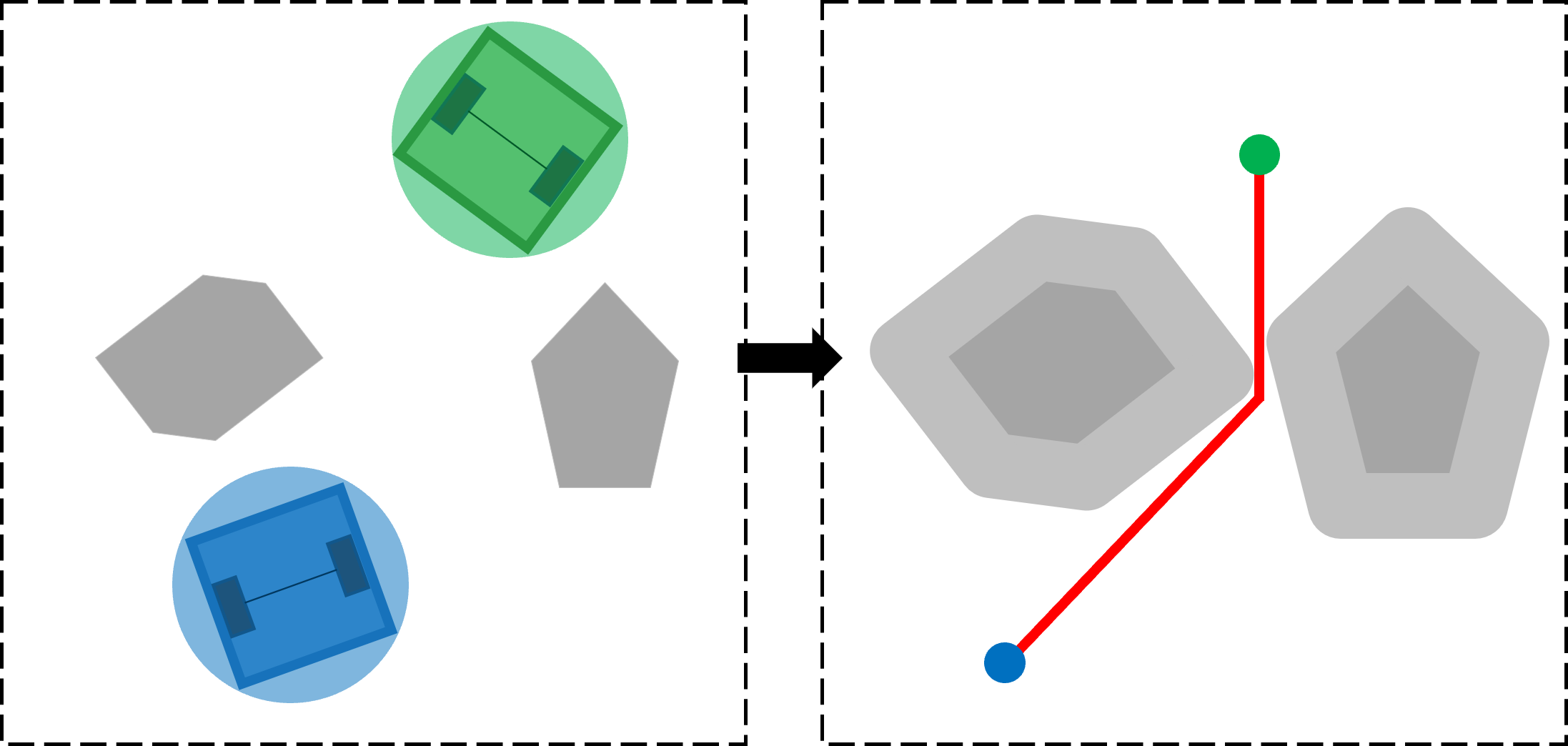} \label{fig:hybrid_dilated}
	}
	\subfigure[The illustration of inverse kinematic calculation of A* path.]{
		\includegraphics[width=.8\linewidth]{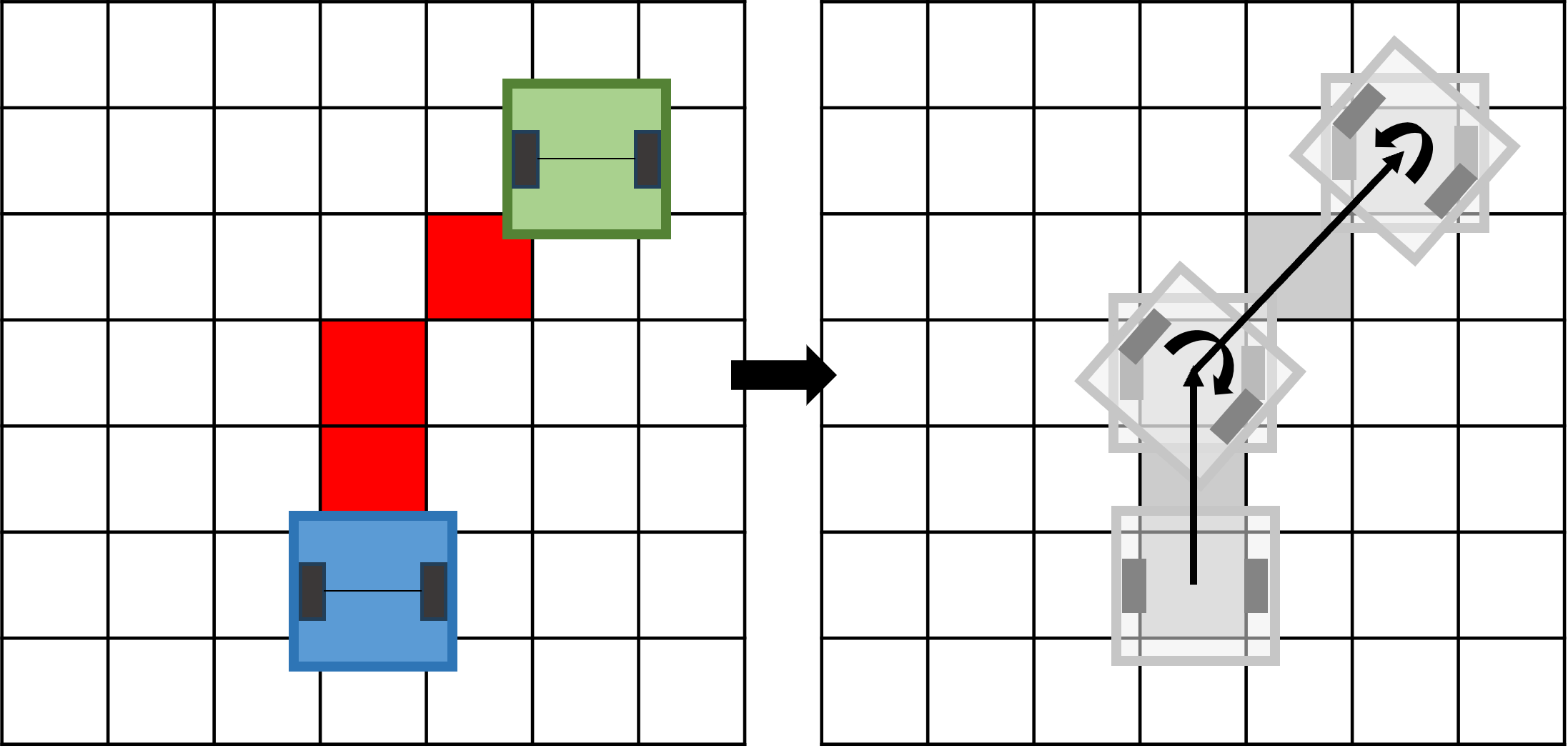} 
		\label{fig:hybrid_target}
 	}
	\caption{In (a), we set the inflated radius to the robot's circumscribed circle radius to ensure collision-free navigation. In (b), the 2D path, initially without consideration of the heading angles, is transformed into a sequence of in-place rotations and straight-line movements.} 
	\label{fig:segmentation_projection}
        \vspace{-10pt}
\end{figure}

\subsubsection{Reward Design}
The reward function is designed to encourage the agent to reach the escape target quickly. The final reward is a normalized linear combination of below three components.
\begin{itemize}
\item \textbf{Intersection-over-Union (IOU) Reward:} The reward is proportional to the IOU area between the robot's boundary and the target boundary.
\item \textbf{Distance Reward:}
This reward provides positive guidance when the robot gets closer to the target.
\item \textbf{Time Consumption Penalty:}
To penalize the agent for longer interaction times, a small negative reward is bounded using a hyperbolic tangent function.
\end{itemize}
Besides, we designed a two-stage curriculum to balance exploration and generalization. Initially, goals were fixed near the entrance to bootstrap learning with dense rewards, followed by randomized goals to encourage the policy to achieve goal-agnostic escape capabilities.
\begin{figure}[t]
	\centering
        \begin{minipage}{0.47\linewidth}
	\subfigure[Discrete action space.]{
	\hfill	\includegraphics[width=.9\linewidth]{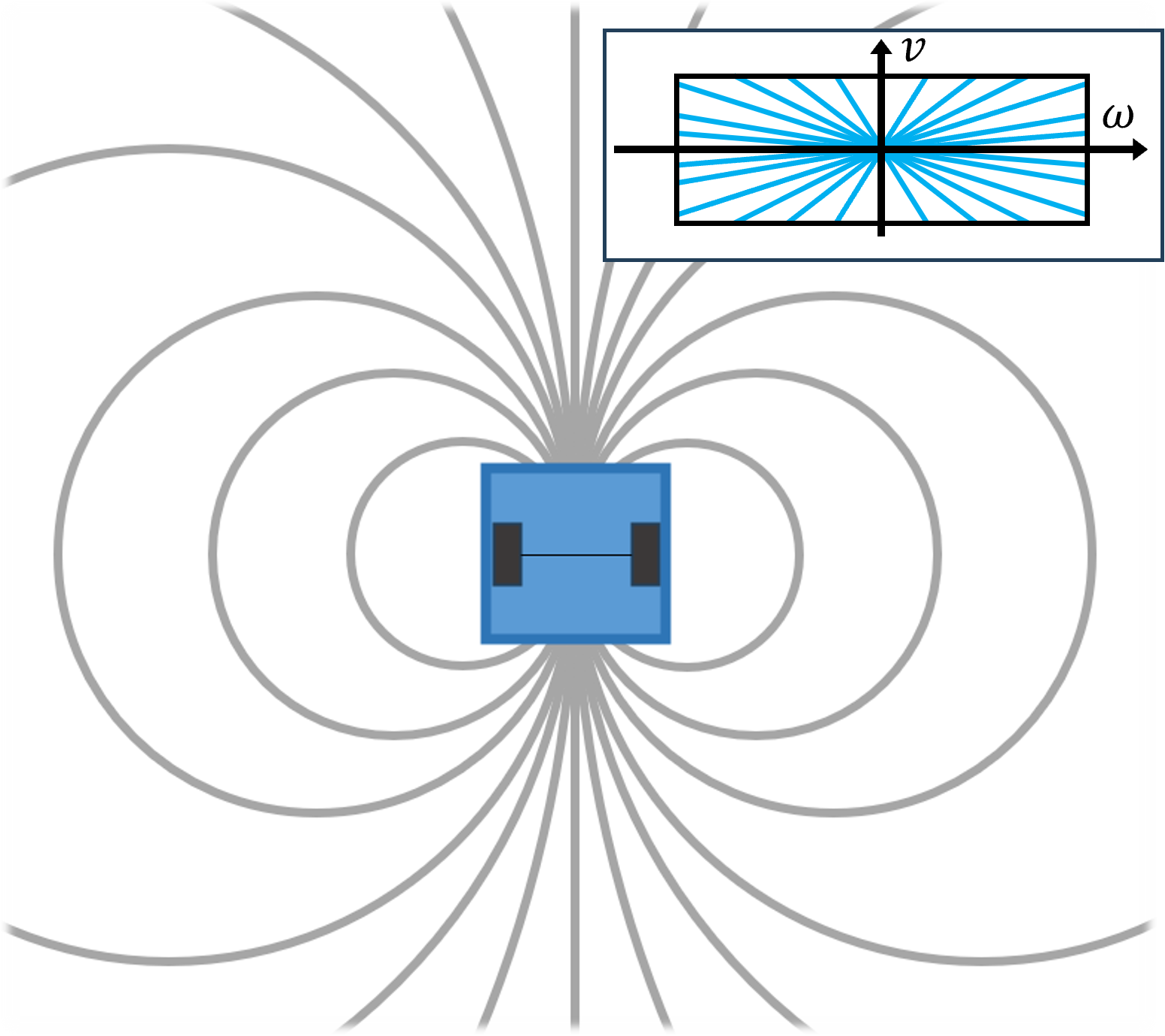} \label{fig:discrete_space}
	}
        \end{minipage}
        \begin{minipage}{0.49\linewidth}
	\subfigure[Action mask calculation.]{
		\includegraphics[width=0.8\linewidth]{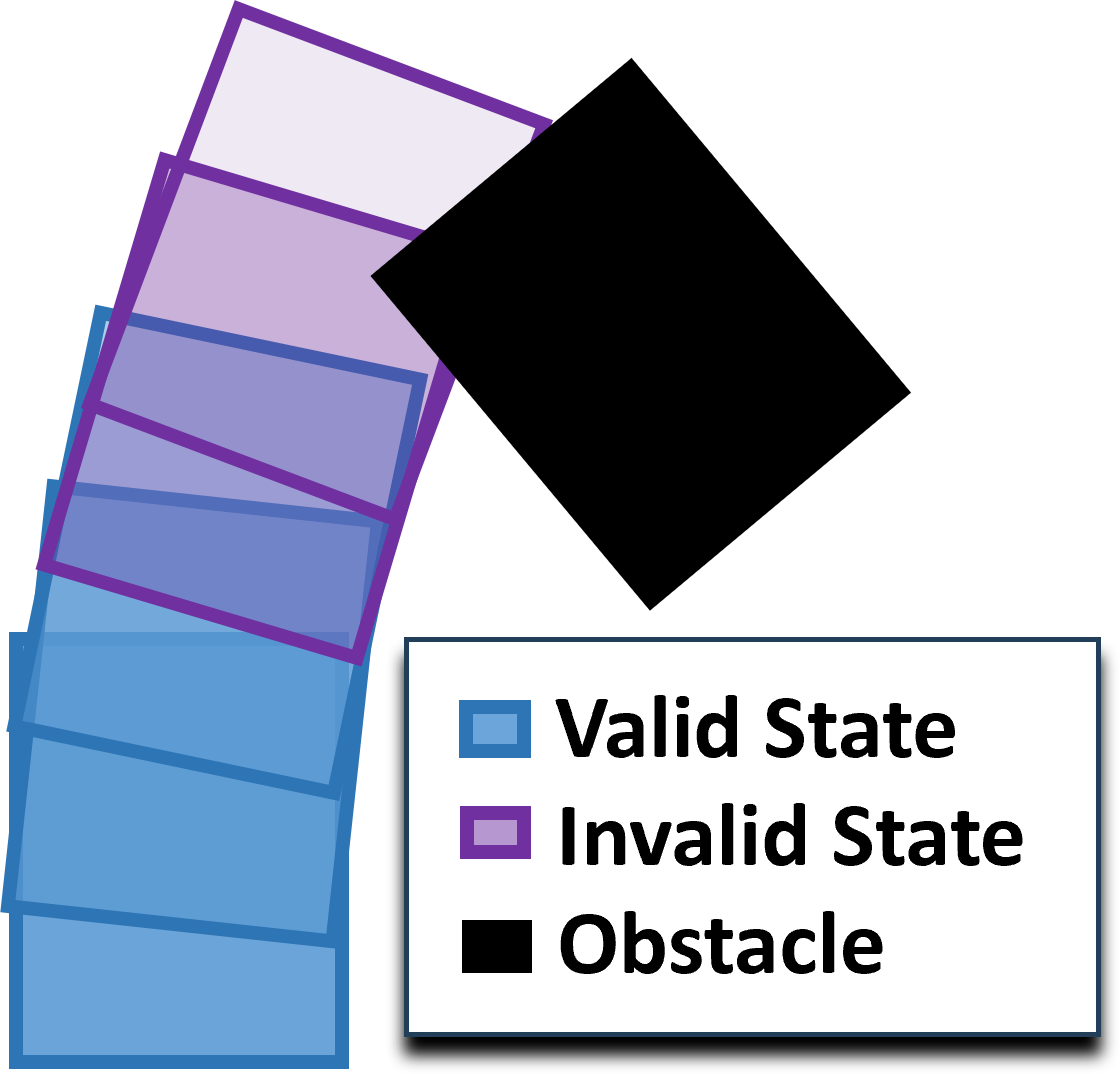} 
		\label{fig:bound_estimate}
 	}
        \end{minipage}
        \subfigure[The Comparison of different intersection.]{
		\includegraphics[width=.7\linewidth]{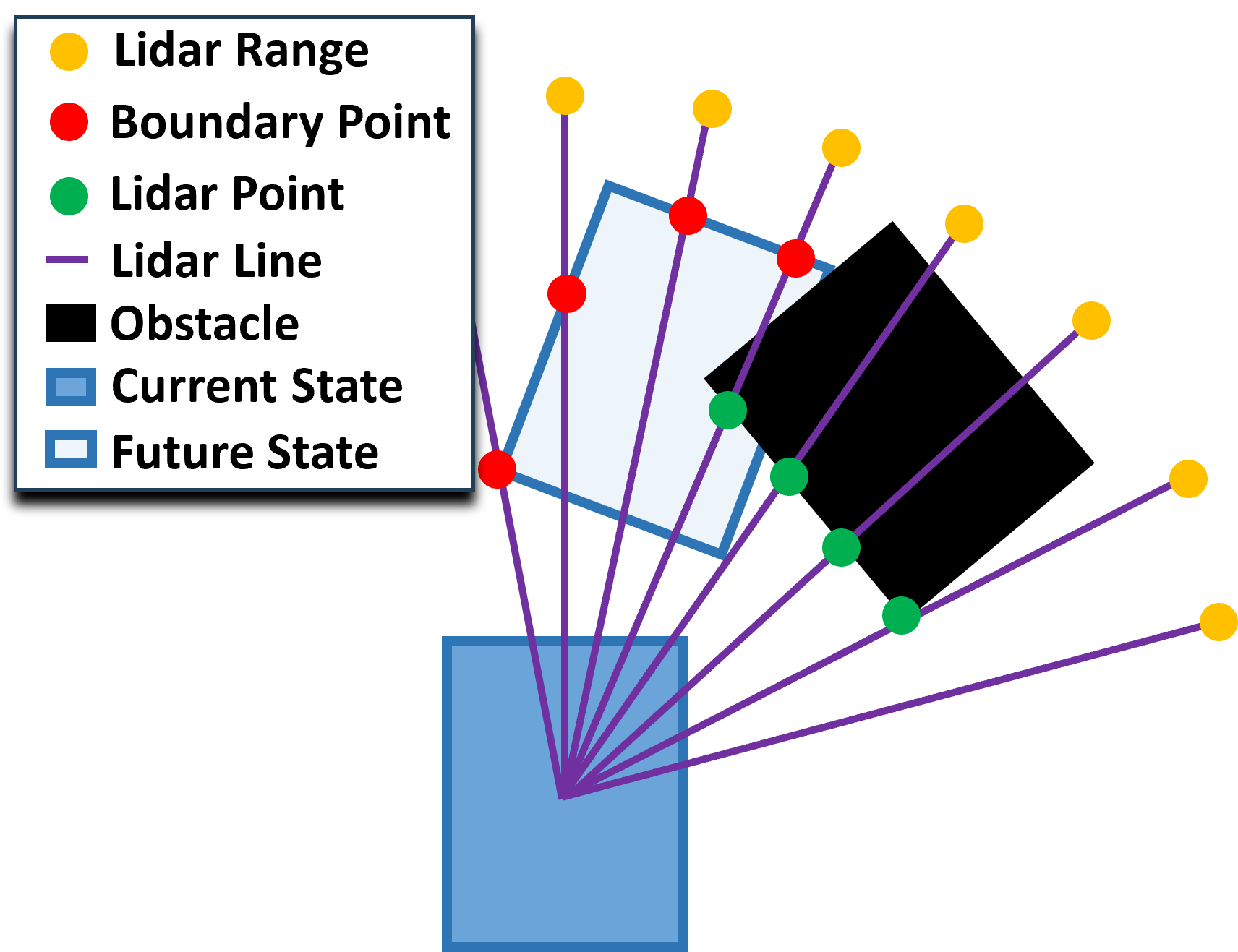} 
		\label{fig:precompute}
 	}
	\caption{In (a), we applied proportional dimensionality reduction to the original action space, restructuring it into a discrete space with uniform turning radius. In (b), The action mask rapidly distinguishes the valid actions and invalid actions. In (c), collision is detected when the distance of the Lidar point is closer than that of the boundary point.} 
	\label{fig:segmentation_projection}
        \vspace{-13pt}
\end{figure}

\begin{figure*}[t]
  \centering
    \subfigure[Constrained space by ROS Nav]{\includegraphics[width=0.23\textwidth]{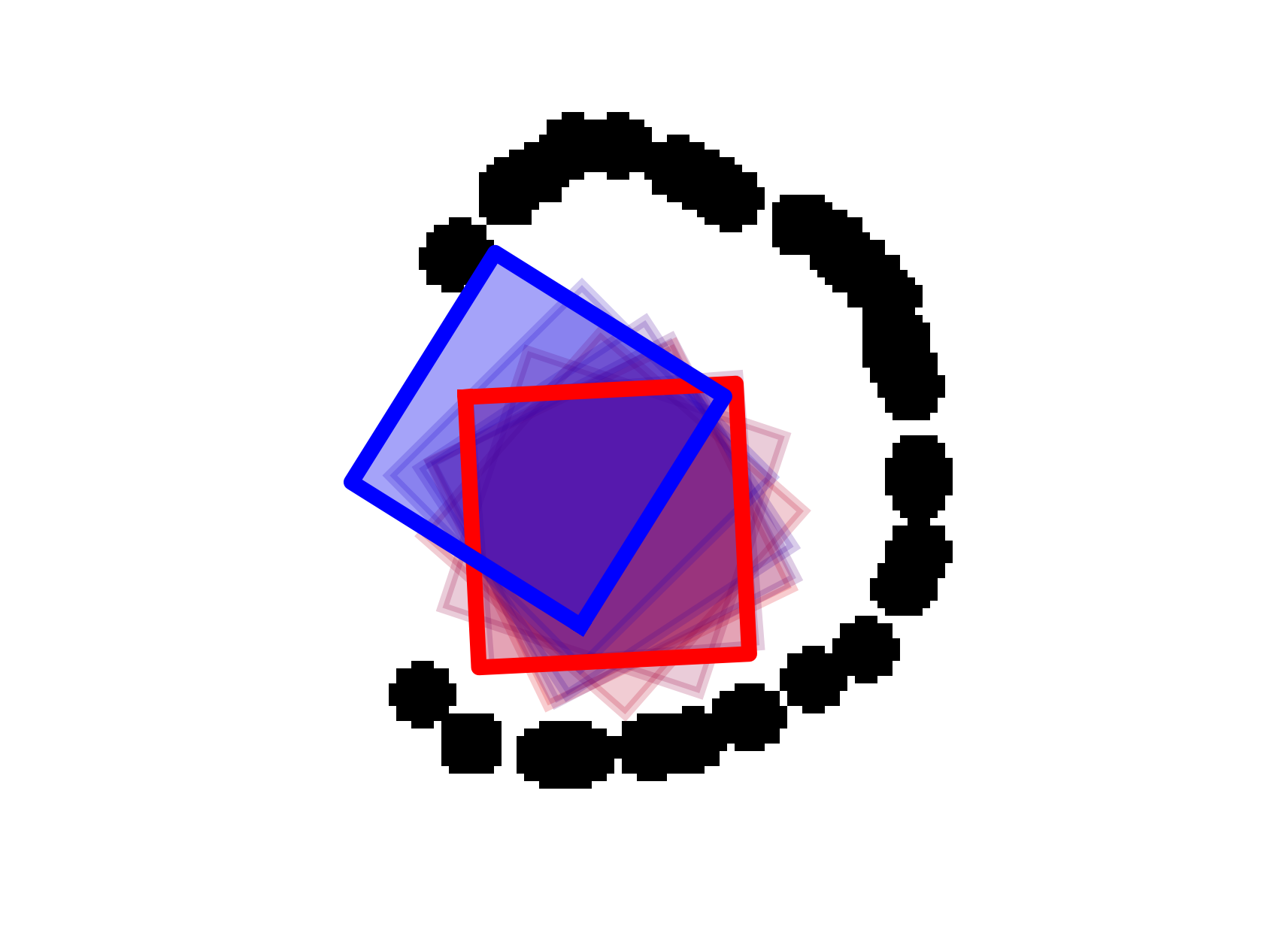}} \hfill
    \subfigure[Sparse obstacles by SAC]{\includegraphics[width=0.23\textwidth]{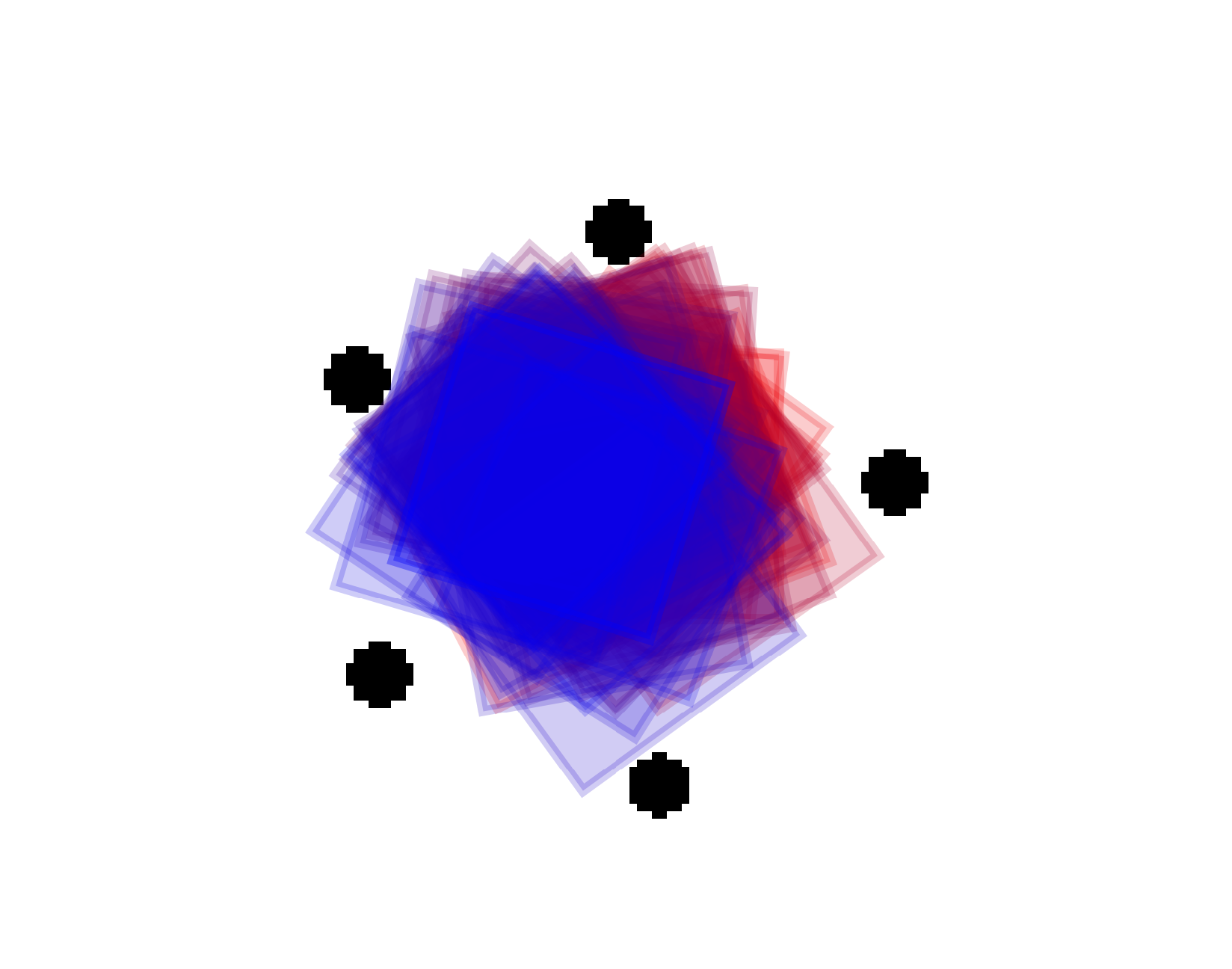}}\hfill
    \subfigure[Long corridor by FastBKRRT]{\includegraphics[width=0.23\textwidth]{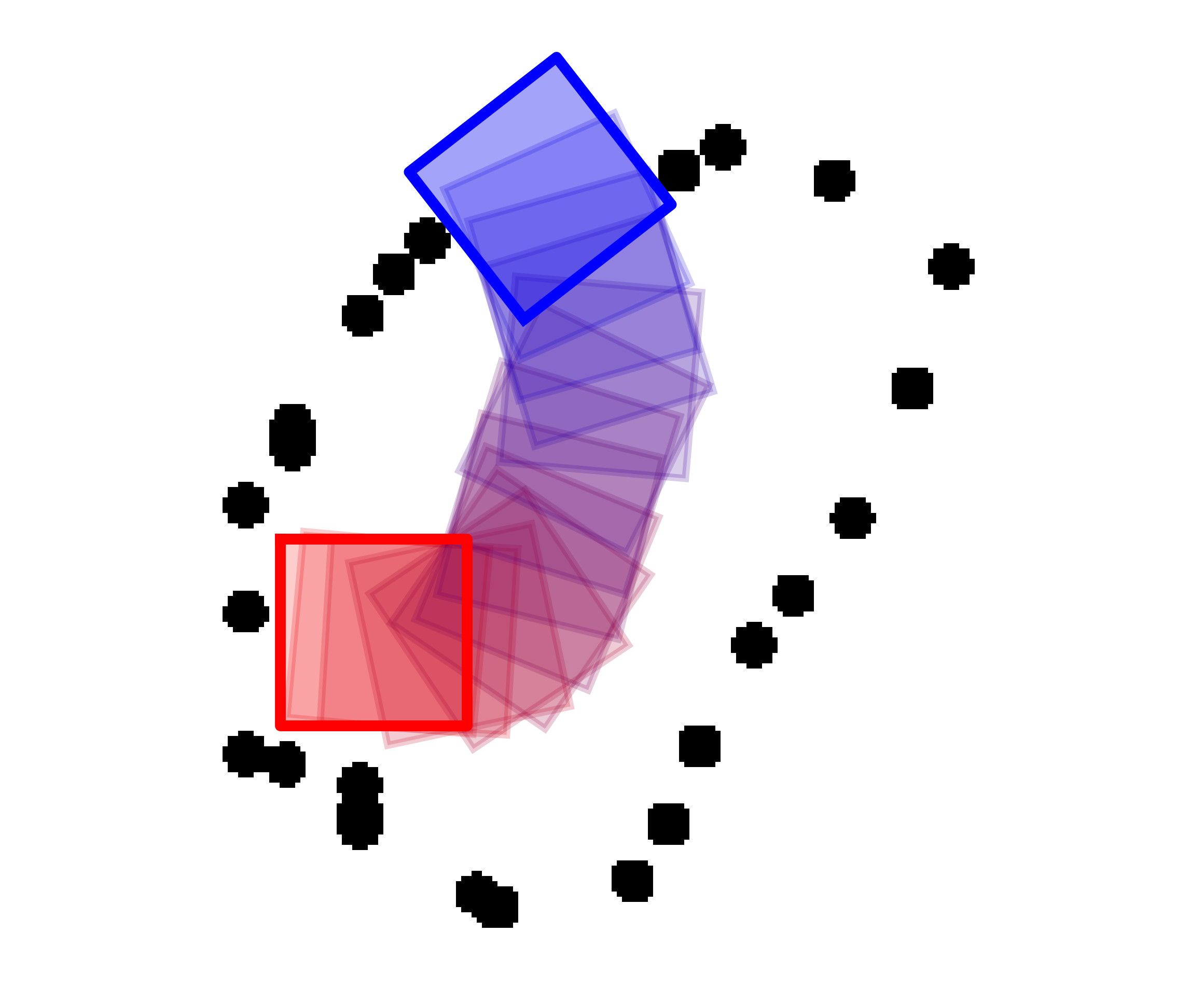}} \hfill
    \subfigure[Long corridor by PPO]{\includegraphics[width=0.23\textwidth]{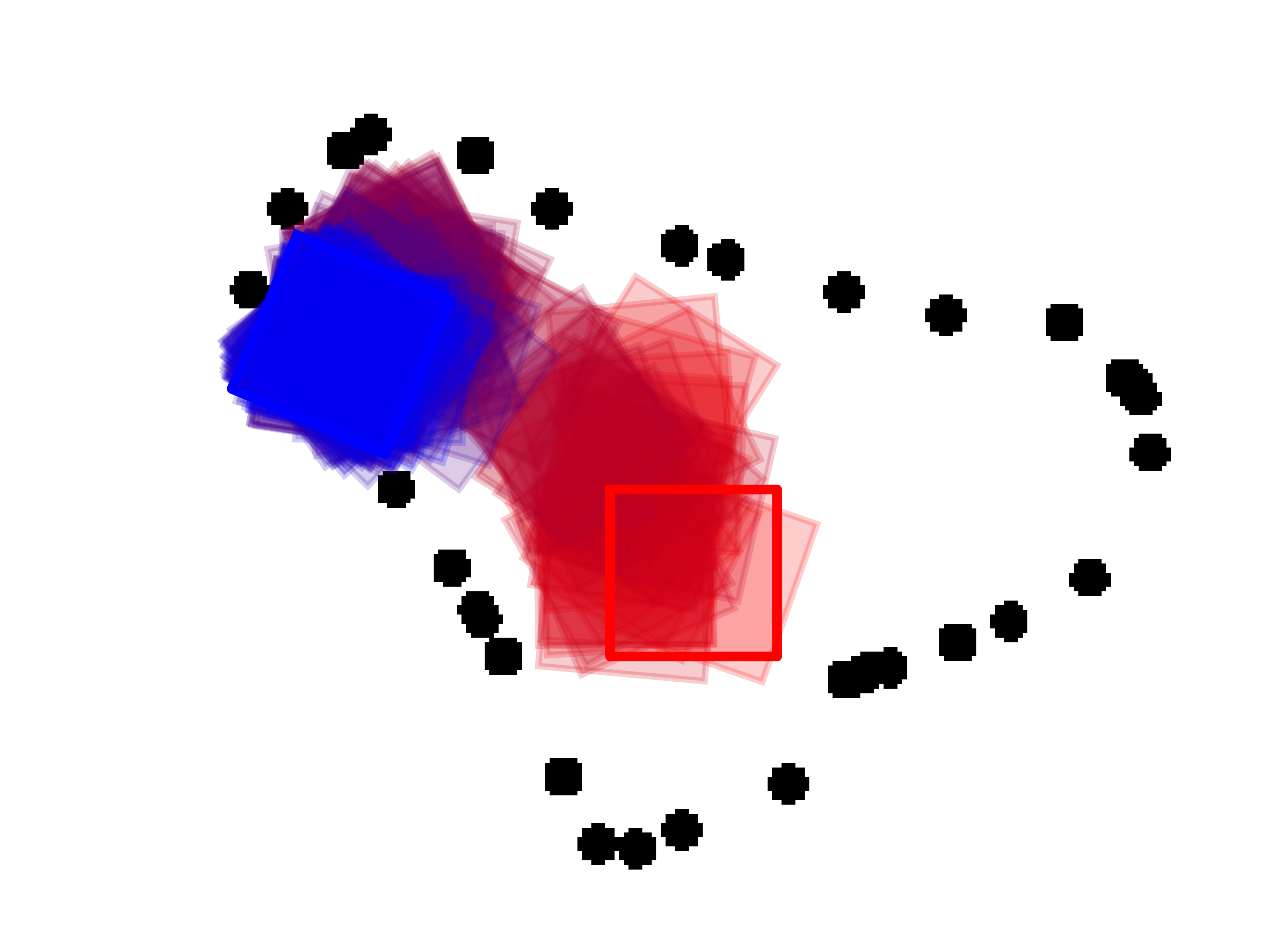}} 
  \newline
  \centering
    \subfigure[Narrow exit by ours]{\includegraphics[width=0.23\textwidth]{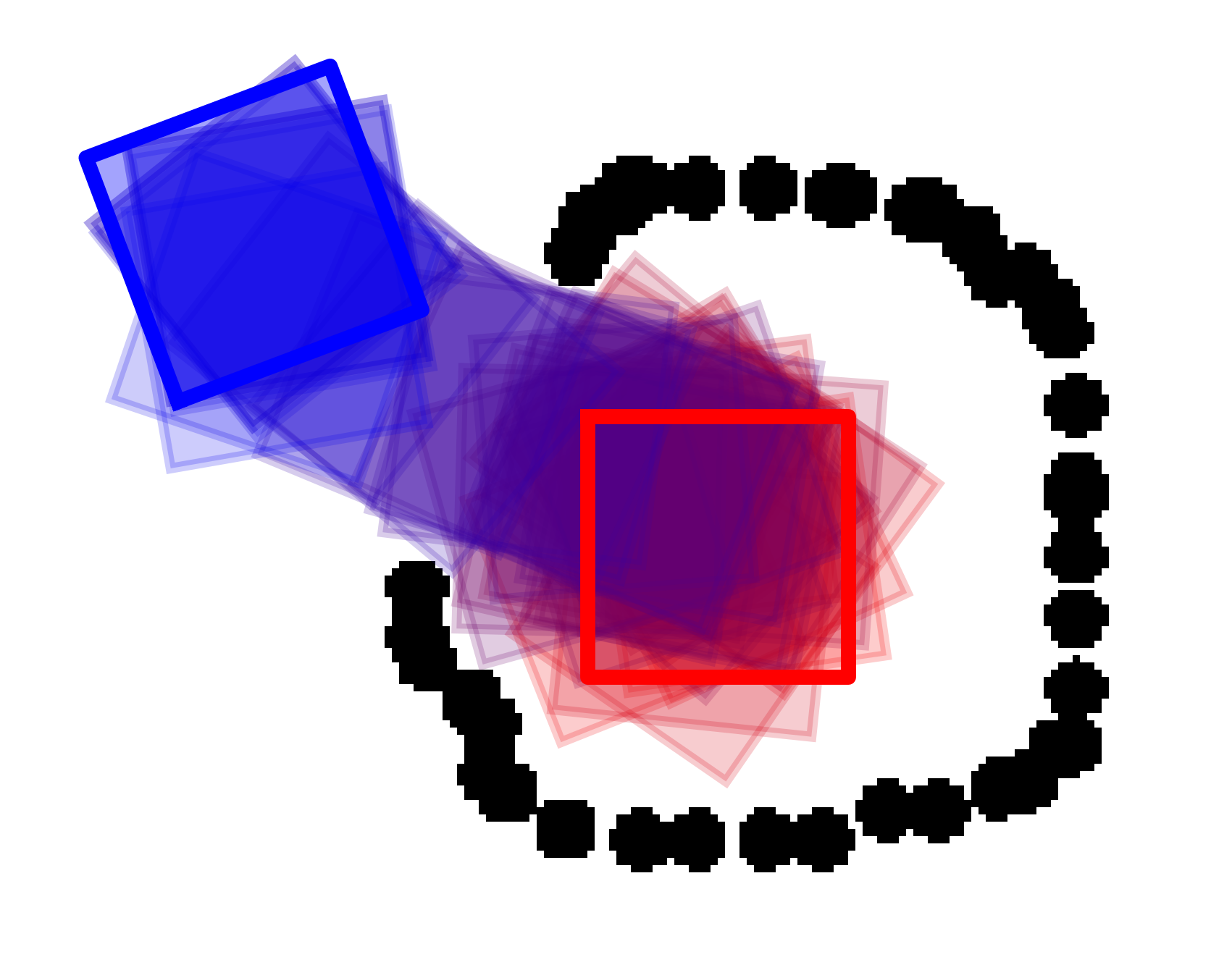}} \hfill
    \subfigure[Constrained space by ours]{\includegraphics[width=0.23\textwidth]{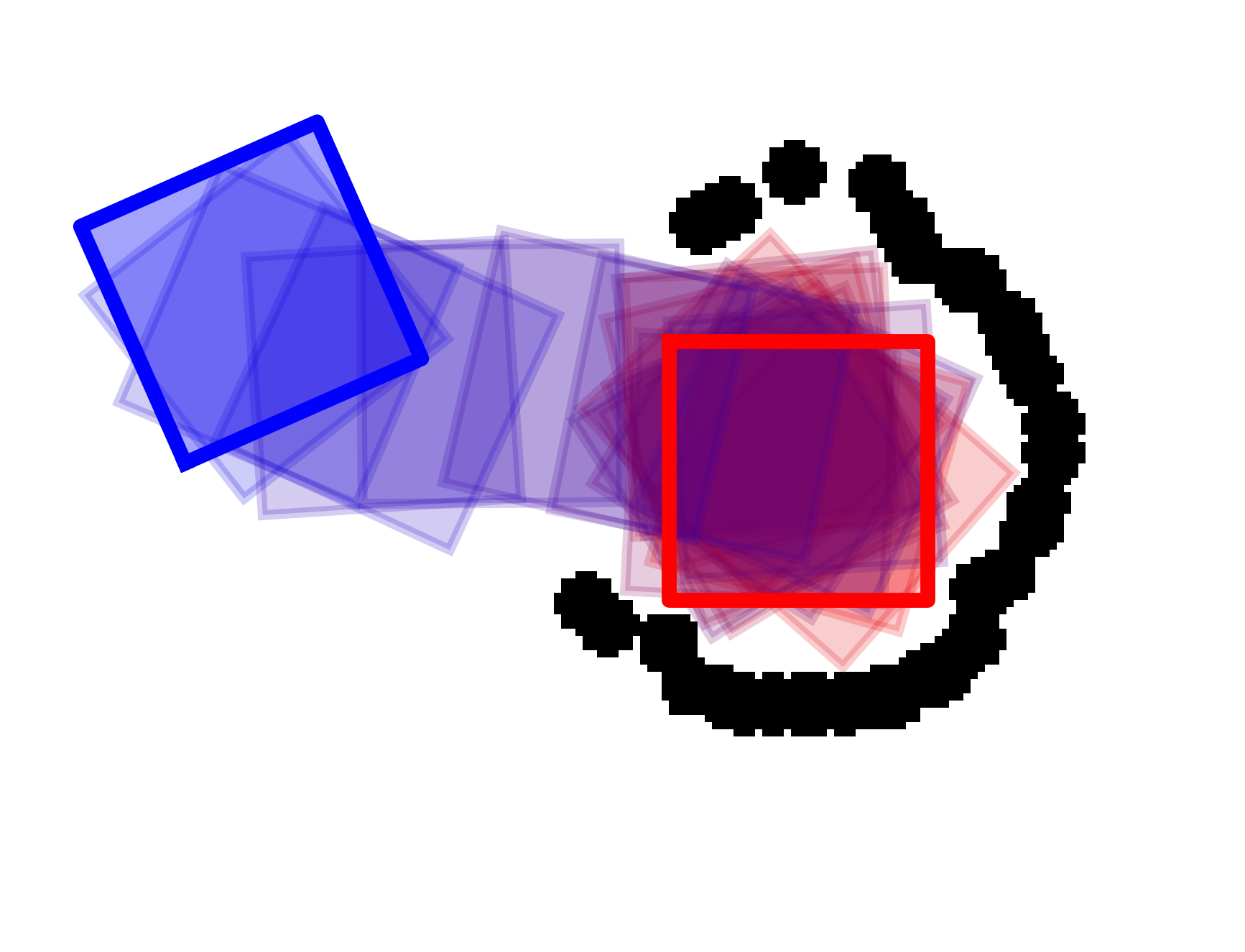}}\hfill
    \subfigure[Sparse obstacles by ours]{\includegraphics[width=0.23\textwidth]{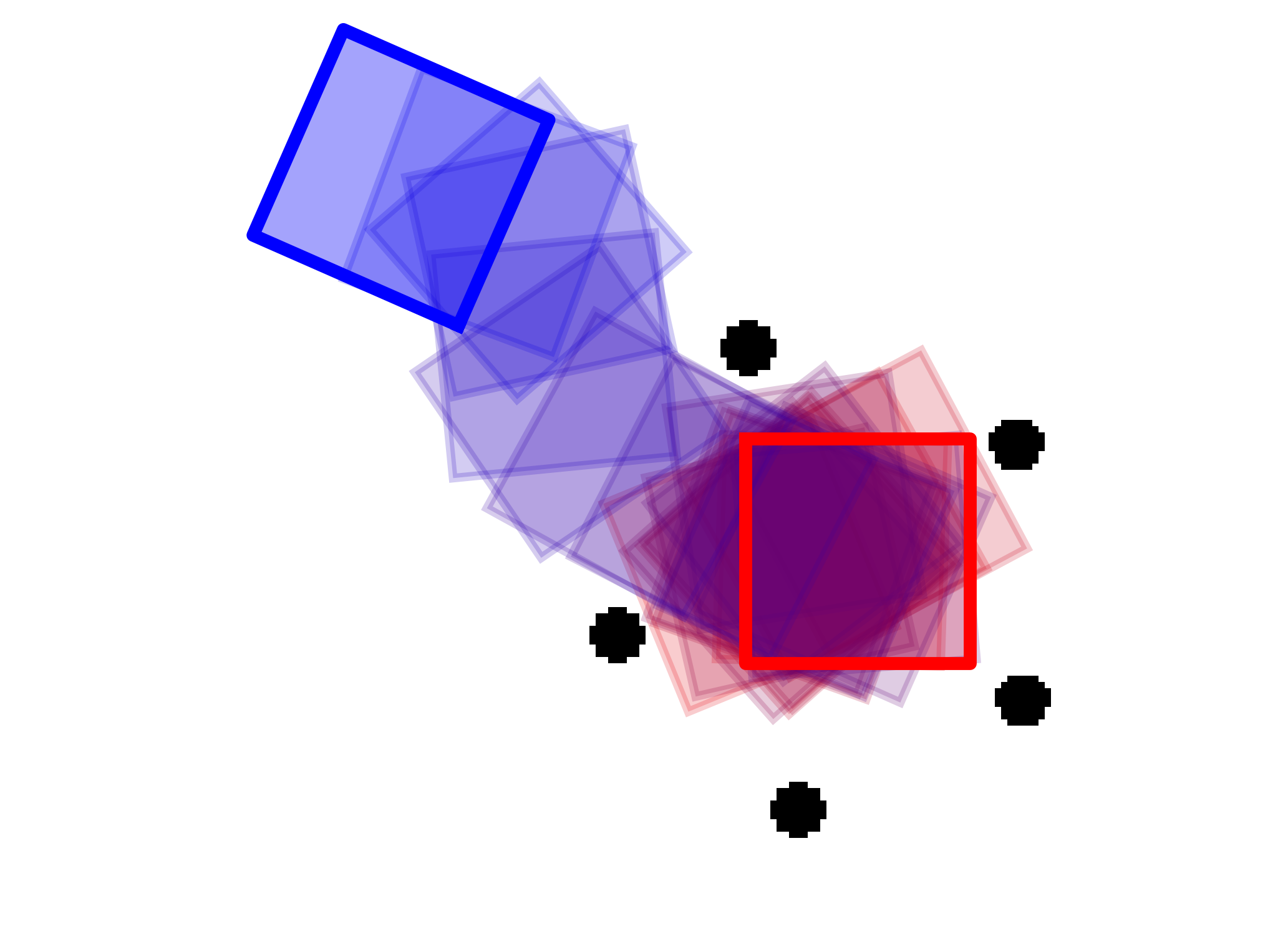}} \hfill
    \subfigure[Long corridor by ours]{\includegraphics[width=0.23\textwidth]{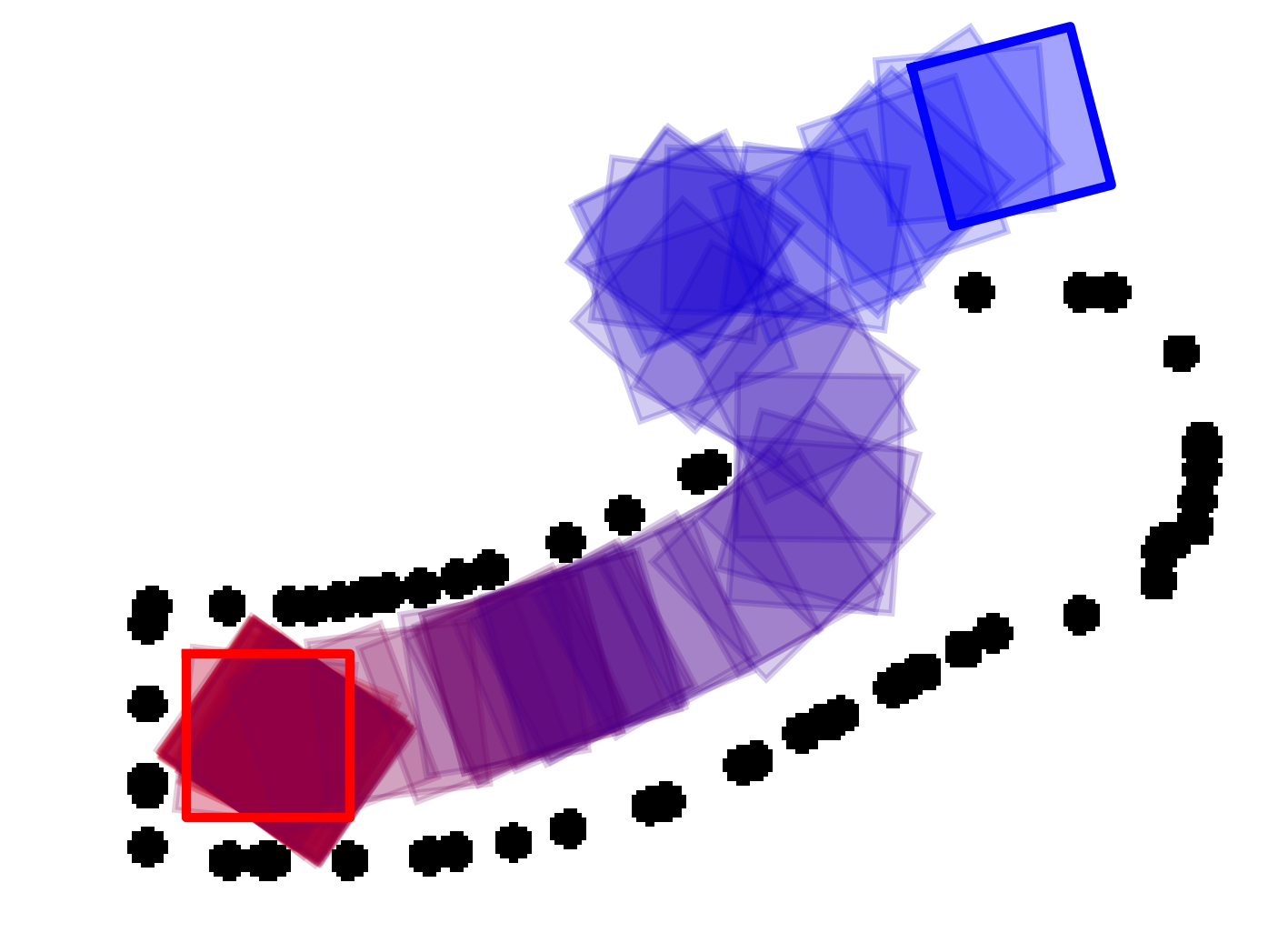}} 
  \caption{The visualization of the escaping process. The starting boundary is highlighted in red, the boundary of the final state is highlighted in blue, and obstacles are filled in black. Red-to-blue gradient rectangles represent the intermediate states explored during the escaping process. In (a)-(d), compared methods fail to escape for collision or endless exploration. In (e)-(h), our approach generates a reasonable trajectory in challenging feature-combination scenarios.}
  \label{fig: case study}
  \vspace{-10pt}
\end{figure*}

\subsection{Hybrid Training Policy}
\label{sec: hybrid}
Navigating in highly constrained spaces presents significant challenges for collision avoidance, leading to sparse rewards during the early stages of reinforcement learning, which can easily cause training to collapse.
To address this issue, we introduce a rule-based algorithm to provide online guidance, mitigating the inefficiency of starting reinforcement learning from scratch.
The hybrid decision-making process combines the RL policy \( \pi_\theta(\cdot \mid s_t) \) and the A* policy \(\pi_{A*}(\cdot \mid s_t) \) to determine the final action taken by the agent.
During each interaction with the environment, the process begins with the exaggerated inflated A* algorithm, which checks for the existence of a feasible path.
Since our robot is square-shaped, when we use the minimum radius for inflation, a path can be found, but it may still result in collisions during execution. Only when the maximum radius is used for dilation does the A* algorithm find a path that is guaranteed to be collision-free, as illustrated in Fig. \ref{fig:hybrid_dilated}.
In this case, the robot's action is calculated using the inverse kinematic model to match the goal pose accurately, as depicted in Fig. \ref{fig:hybrid_target}.
This method guides the robot through relatively simple environments during the early stages of training.
The action from A* policy \(\pi_{A*}(\cdot \mid s_t) \) is only utilized when the path is feasible. This ensures that the robot can still navigate through the RL policy \( \pi_\theta(\cdot \mid s_t) \) in complex environments where traditional planning methods may fail. Additionally, it’s important to note that the A* algorithm is only used during training to provide online guidance. During testing, we rely solely on the RL policy as the executor.

\subsection{Efficient Action Mask}
\label{sec: action_mask}
The action mask mechanism, which identifies and excludes invalid actions, is highly beneficial for reinforcement learning training \cite{huang20222closer}. 
However, for differential-drive robots, this requires checking whether all discretized actions $(\omega_i, v_i)$ in the 2D action space would lead to collisions, which is computationally inefficient. 
Some works that apply action mask in autonomous driving vehicles\cite{jiang2024hope}, quadrotors\cite{stolz2025excluding} and manipulators\cite{wu2024efficient} are not suitable for our vacuum robots.
To address this, we designed a dimensionality reduction method for the action space tailored to differential-drive robots, enabling more efficient implementation of the action mask.
Specifically, when \(\lvert \frac{v_i}{\omega_i} \rvert\) remains constant, the robot rotates around a fixed point. 
Then calculating the maximum collision-free step size (denoted as $\Delta S_i$) for a given action $(\omega_i, v_i)$ is equivalent to obtain the $\Delta S_j = \Delta S_i/k$ for any action $(\omega_j, v_j)$ if $\omega_j=k\omega_i$ and $v_j=kv_i$. 
The proportional dimensionality reduction results in a contour-like action space, 
as shown in Fig. \ref{fig:discrete_space}. The discrete action space $A_{\text{discrete}}$ includes the actions at 42 turning directions and is defined below:
\begin{equation}
A_{\text{discrete}} = \underbrace{\{(\pm\omega_{\max}, 0)\}}_{\text{In-place rotation}} 
\cup 
\bigcup_{i=0}^{9} 
\underbrace{\{(\pm\omega_i)\} \times \{(\pm v_i)\}}_{\text{Constant-radius turning}}
\end{equation}
where:
\begin{equation}
\begin{cases}
\omega_i = \min\left(\omega_{\max}, \frac{v_{\max}}{r_i} \right) \\
v_i = \min\left(v_{\max}, \omega_{\max} \cdot r_i \right) \\
r_i = 0.01 \times 2^i (m)\\
i = 0,1,...,9.
\end{cases}
\end{equation}
Here, $\omega_{\max}$ and $v_{\max}$ represent the robot’s maximum angular and linear velocities, respectively. The turning radius $r_i$ follows an exponential sequence, ensuring a diverse range of motion from sharp turns to straight line. The constraints on $\omega_i$ and $v_i$ guarantee that each action remains within the robot’s physical limitations while maintaining kinematic feasibility.

We have reduced the 2D planar space to a 1D linear space, which enables the implementation of an efficient action mask mechanism.
Inspired by \cite{jiang2024hope}, we develop a precompute action mask to identify the invalid states and filter out safe action, as shown in Fig. \ref{fig:bound_estimate}. The core insight behind this approach is that we can pre-generate the next-state bounding boxes for all possible discretized actions and calculate the boundary points in the directions of laser scan, as the red points shown in Fig. \ref{fig:precompute}. Thus, when the new Lidar point cloud data (green points) is received, the validity of actions can be obtained by directly comparing the Lidar points and the boundary points. Compared with common collision checks which usually serve as post-processing, this approach can also provide a prior feature of the distribution of valid actions for our neural network, which helps robots focus on learning safe actions and inference in real-time.

\label{sec:reshape}

\begin{table*}[ht]
  \renewcommand\arraystretch{1.2}
  \centering
  \caption{Total Success Rate of Different Approaches For Various Challenging Features.}
  \begin{tabular}{ccccccccccccccccc} 
    \toprule
    \multirow{2.5}*{\makecell[c]{N}} &
    \multirow{2.5}*{\makecell[c]{C}} &
    \multirow{2.5}*{\makecell[c]{S}} &
    \multirow{2.5}*{\makecell[c]{L}} &

    \multicolumn{4}{c}{\centering ROS Navigation} &
    \multicolumn{4}{c}{\centering FastBKRRT} &
    \multicolumn{5}{c}{\centering RL Network } \\
    \cmidrule(r){5-8} \cmidrule(r){9-12} \cmidrule(r){13-17}
    & & & &
    \multicolumn{1}{c}{\centering 18cm} &
    \multicolumn{1}{c}{\centering 20cm} &
    \multicolumn{1}{c}{\centering 23cm} &
    \multicolumn{1}{c}{\centering 25cm} &
    \multicolumn{1}{c}{\centering 18cm} &
    \multicolumn{1}{c}{\centering 20cm} &
    \multicolumn{1}{c}{\centering 23cm} &
    \multicolumn{1}{c}{\centering 25cm} &
    \multicolumn{1}{c}{\centering SAC} & 
    \multicolumn{1}{c}{\centering PPO} &
    \multicolumn{1}{c}{\centering SAC-AM} & 
    \multicolumn{1}{c}{\centering SAC-HY} &
    \multicolumn{1}{c}{\centering Ours} \\
    \midrule
    $\checkmark$ & $\checkmark$ & $\checkmark$ & $\checkmark$ & $0.000$ & $0.125$ & $0.283$ & $0.239$ & $0.194$ & $0.364$ & $0.289$ & $0.233$ & $0.200$ & $0.170$ & $0.235$ & $0.290$ & $\textbf{0.751}$ \\
    $\checkmark$ & $\checkmark$ & $\checkmark$ & $ $ & $0.000$ & $0.211$ & $0.361$ & $0.239$ & $0.225$ & $0.386$ & $0.367$ & $0.214$ & $0.360$ & $0.190$ & $0.410$ & $0.580$ & $\textbf{0.780}$ \\
    $\checkmark$ & $\checkmark$ & $ $ & $ $ & $0.003$ & $0.214$ & $0.356$ & $0.222$ & $0.192$ & $0.372$ & $0.353$ & $0.211$ & $0.385$ & $0.310$ & $0.550$ & $0.895$ & $\textbf{0.945}$ \\
    $\checkmark$ & $ $ & $ $ & $ $  & $0.000$ & $0.261$ & $0.469$ & $0.231$ & $0.272$ & $0.583$ & $0.439$ & $0.214$ & $0.420$ & $0.465$ & $0.625$ & $0.895$ & $\textbf{0.995}$\\
     $ $ & $ $ & $ $ & $ $ & $0.003$ & $0.344$ & $0.808$ & $0.753$ & $0.414$ & $0.722$ & $0.800$ & $0.717$ & $0.500$ & $0.475$ & $0.650$ & $0.965$ & $ \textbf{0.990} $\\
    \bottomrule
  \end{tabular}
  \label{tab:success_rate}
\end{table*}

\begin{table*}[ht]
  \renewcommand\arraystretch{1.2}
  \centering
  \begin{threeparttable}
  \caption{Planning and Control Success Rate of Traditional Algorithms.}
  \begin{tabular}{ccccccccccccccc} 
    \toprule
    \multirow{2.5}*{\makecell[c]{Success rate \\on test set}} &
    \multicolumn{4}{c}{\centering ROS Navigation} &
    \multicolumn{4}{c}{\centering FastBKRRT} &
    \multicolumn{5}{c}{\centering RL Network } \\
    \cmidrule(r){2-5} \cmidrule(r){6-9} \cmidrule(r){10-14}
    &
    \multicolumn{1}{c}{\centering 18cm} &
    \multicolumn{1}{c}{\centering 20cm} &
    \multicolumn{1}{c}{\centering 23cm} &
    \multicolumn{1}{c}{\centering 25cm} &
    \multicolumn{1}{c}{\centering 18cm} &
    \multicolumn{1}{c}{\centering 20cm} &
    \multicolumn{1}{c}{\centering 23cm} &
    \multicolumn{1}{c}{\centering 25cm} &
    \multicolumn{1}{c}{\centering SAC} & 
    \multicolumn{1}{c}{\centering PPO} &
    \multicolumn{1}{c}{\centering SAC-AM} & 
    \multicolumn{1}{c}{\centering SAC-HY} &
    \multicolumn{1}{c}{\centering Ours} \\
    \midrule
    Planning succ. rate & $1.000$ & $0.986$ & $0.761$ & $0.458$ & $0.892$ & $0.881$ & $0.678$ & $0.372$ & $\backslash$ & $\backslash$ & $\backslash$ & $\backslash$ & $\backslash$ \\
    Control succ. rate& $0.003$ & $0.206$ & $0.635$ & $0.908$ & $0.265$ & $0.530$ & $0.684$ & $0.851$ & $\backslash$ & $\backslash$ & $\backslash$ & $\backslash$ & $\backslash$ \\
    Total succ. rate& $0.003$ & $0.203$ & $0.483$ & $0.372$ & $0.236$ & $0.467$ & $0.464$ & $0.317$ & $0.400$ & $0.292$ & $0.469$ & $0.564$ & $\textbf{0.758}$ \\
    \bottomrule
  \end{tabular}
\vspace{0.2cm}


\centering
N, C, S, L: denote four challenging features respectively for limitation of table size. 

  \label{tab:planning_control}
  \end{threeparttable}
\end{table*}

\section{Experiments}

\subsection{Experimental Setup}
The reinforcement learning policy is trained in a customized 2D simulator developed based on the open-source simulator Tactics2D \cite{li2023tactics2d}. Apart from our lightweight simulator, the automated generation of diverse narrow scenarios is crucial for RL training. However, randomly generated scenarios do not guarantee the existence of an escaping path. Therefore, our approach is to first generate a feasible path, and then construct obstacles around this path.
In practice, the robot is moved randomly and the envelope of its trajectory is used to generate a connected occupied space. Then obstacles are randomly placed out of this occupied space.
We utilize our random scenario generator to create a training set consisting of 1,800 scenarios and a test set with 360 scenarios across the four challenging features below.
\begin{itemize}
    \item \textbf{Narrow Exit}: Scenarios with the open size smaller than the circumscribed circle radius of the robot.
    \item \textbf{Constrained Space}: Determined by the distance between obstacles and the robot's contour.
    \item \textbf{Sparse Obstacles}: If the distance between obstacles is not large enough, the gaps may mislead robot learning.
    \item \textbf{Long Corridor}: Long-distance navigation requires more unknown exploration.
\end{itemize}

In each episode, the simulator imports a scenario and randomly initializes feasible start and goal positions. 
A total of 20,000 training episodes were conducted, with 1,000 trials during testing.
Besides, the training process was performed on an NVIDIA GeForce RTX 4070 GPU and Intel i7-14700KF CPU. 

\subsection{Simulation Results}

To evaluate the performance of our proposed method, we conducted extensive simulation experiments in our 2D simulator. 
We compared our approach with the following baseline methods:
\begin{itemize}
    \item \textbf{ROS Navigation} (ROS Nav)\cite{zheng2021ros}: A widely-used 2D navigation stack improved on A* algorithm \cite{hart1968formal}, which takes in information from odometry, sensor streams, and a goal pose and outputs safe velocity commands to navigate in complex environments.
    \item \textbf{FastBKRRT}\cite{peng2021towards}: It's an optimized algorithm based on RRT* \cite{noreen2016optimal} algorithm, specifically designed for environments with narrow passages and dense obstacles. While this work is originally designed for Ackermann-steering vehicles, we adapted origin Ackermann kinematics to our differential-drive pure pursuit controller \cite{samuel2016review}.
    \item \textbf{SAC}\cite{haarnoja2018soft} and \textbf{PPO}\cite{schulman2017proximal}: SAC is a maximum entropy off-policy reinforcement learning method that maximizes cumulative reward and policy entropy. PPO is a policy-gradient-based on-policy method performing gradient ascent within a trust region. We explored naive approaches in our work.
    \item \textbf{SAC-AM} and \textbf{SAC-HY}: We apply action mask mechanism and hybrid training policy to SAC network respectively to prove that only using one component falls short in addressing the challenges of escaping. We employ the same SAC network as ours for fair comparison.
\end{itemize}

\begin{figure}[t]
	\centering
	\includegraphics[width=1.0\linewidth]{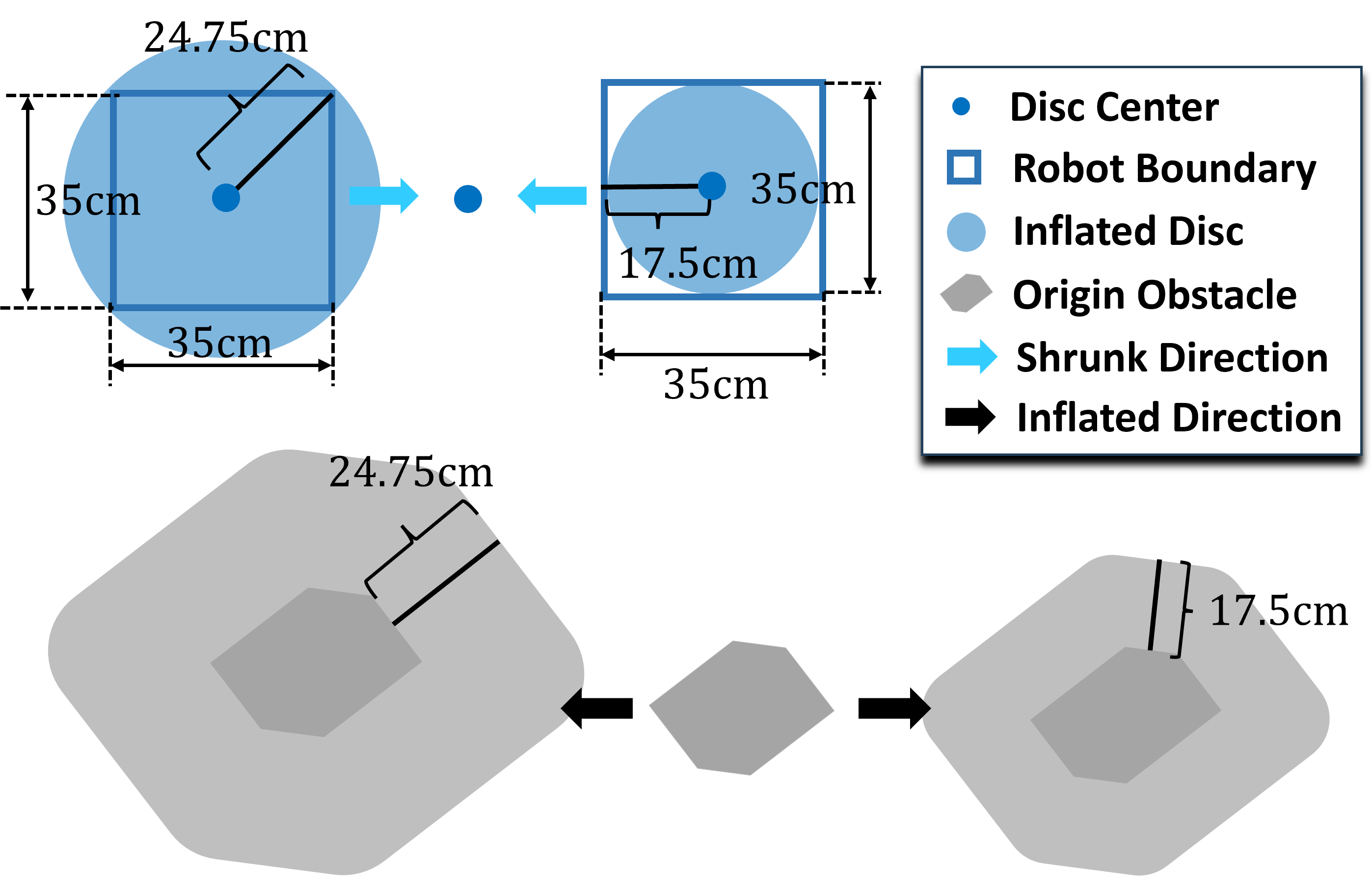} 
	\caption{The demonstration of the inflation radius concept for collision checking. A square robot covered by a disc is simplified to a disc center by expanding origin obstacles with an equivalent radius, converting collision detection into point-in-inflated-obstacle checks. The figure also visualizes how dynamic inflation radius adjustments alter both robot representation and obstacle boundaries.} 
    \label{fig:Inflation}
        \vspace{-10pt}
\end{figure}

We designed four feature-combination scenarios to simulate diverse real-world escape environments. 
As shown in Table \ref{tab:success_rate}, the features include narrow exits, constrained spaces, sparse obstacles, and long corridors. 
A checkmark indicates that a scenario possesses the feature. 
Scenarios without these four features are classified as simple. 
Each kind of scenario was randomly generated 200 times.

Fig. \ref{fig: case study} visualizes the robot’s escape trajectories across different scenarios.
In addition to the visual qualitative results, we present quantitative comparisons between traditional algorithms and our method under varying inflated radius.
Fig. \ref{fig:Inflation} visualize the relationship between the inflated radius and robot's size.
The inflated radius is sampled between the robot's width and half of its diagonal. 
As shown in Table \ref{tab:success_rate}, our approach consistently outperforms the baselines in total success rates of escaping task, especially in scenarios with narrow exits. 
While ROS Navigation and FastBKRRT perform well in less constrained environments, their performance deteriorates significantly as the scenario becomes more cluttered and narrow, as shown in Fig. \ref{fig: case study}(a)(c). 

To explain the inefficiencies of traditional methods in escape tasks, we conducted a comparison experiment in our test set. We decompose the total success rate into planning success rate and control success rates. 
The planning success rate measures the proportion of successful planning paths regardless of kinematic constraints for rule-based approaches, while the control success rate tracks the collision-free execution of the planned path.
As shown in Table \ref{tab:planning_control}, traditional algorithms are highly sensitive to the inflated radius. 
A larger inflated radius reduces free space and lowers the planning success rate but increases the control success rate due to greater safety margins. 
ROS Navigation typically generates better paths than FastBKRRT, but FastBKRRT achieves a higher average total success rate, indicating that ROS Navigation’s paths are often close to obstacles, making them more prone to control errors and collisions. 
This also explains why ROS Navigation achieves a nearly 100\% planning success rate but a nearly 0\% control success rate when the inflated radius is half of the robot’s width. 
In contrast, our end-to-end architecture achieves a high total success rate across challenging environments, balancing planning and control effectively where traditional decoupled approaches struggle.

We also explored naive SAC and PPO for comparison, and experiments revealed that both of them perform poorly in escaping task, as shown in Fig. \ref{fig: case study}(b)(d). 
Moreover, we conduct comprehensive ablation experiments, as shown in Table \ref{tab:success_rate} and Table \ref{tab:planning_control}. 
Against cold-start RL from scratch without prior guidance, our hybrid training policy reduces fail rates by 61.6\% in complex environments.
Compared to sampling without action mask, our dimensionality reduction masking explicitly incorporates constraints, achieving an 34.4\% higher success rate in narrow space.
The significant drop in performance of success rate in the absence of either the customized action masking mechanism or hybrid training policy demonstrates the critical importance of both components for effective escaping.

\subsection{Real-world Experiments}

\begin{figure}[t]
  \vspace{10pt}
  \centering
    \subfigure[Narrow exit]{\includegraphics[width=0.235\textwidth]{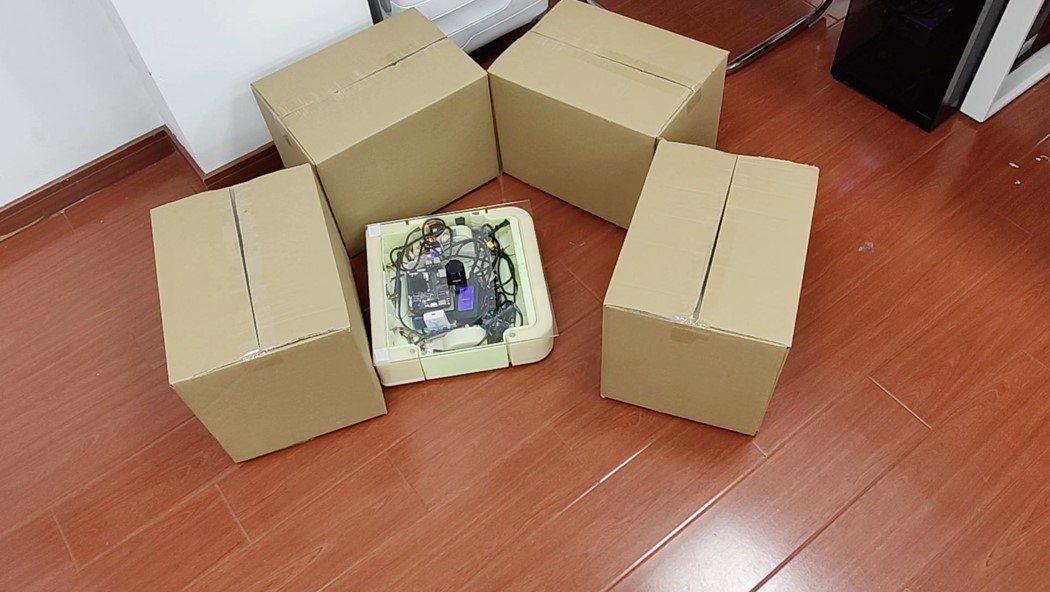}}
    \hfill
    \subfigure[Long corridor]{\includegraphics[width=0.235\textwidth]{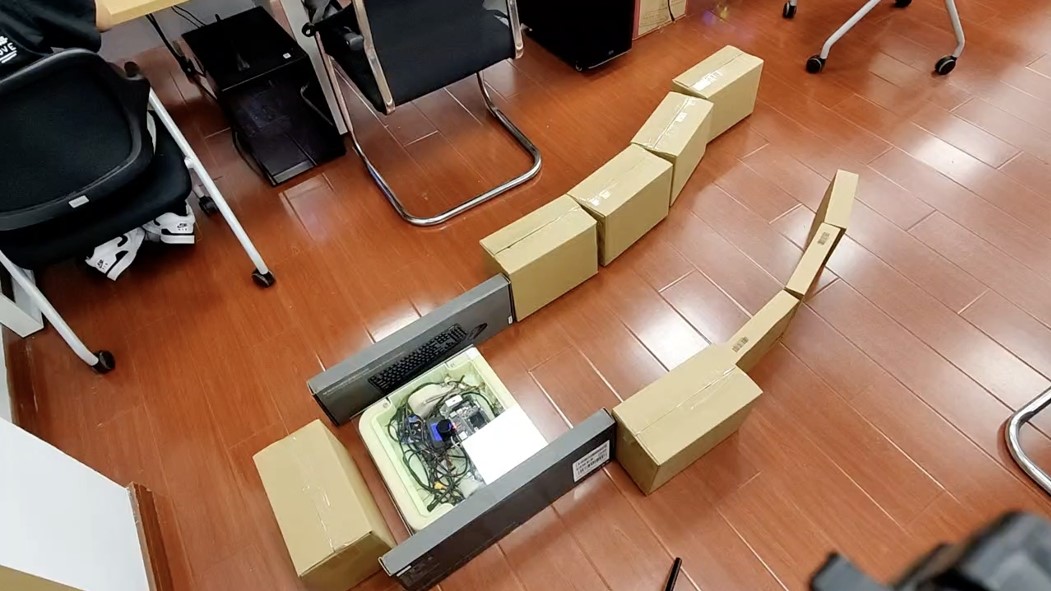}}
    \newline
    \subfigure[Chair]{\includegraphics[width=0.235\textwidth]{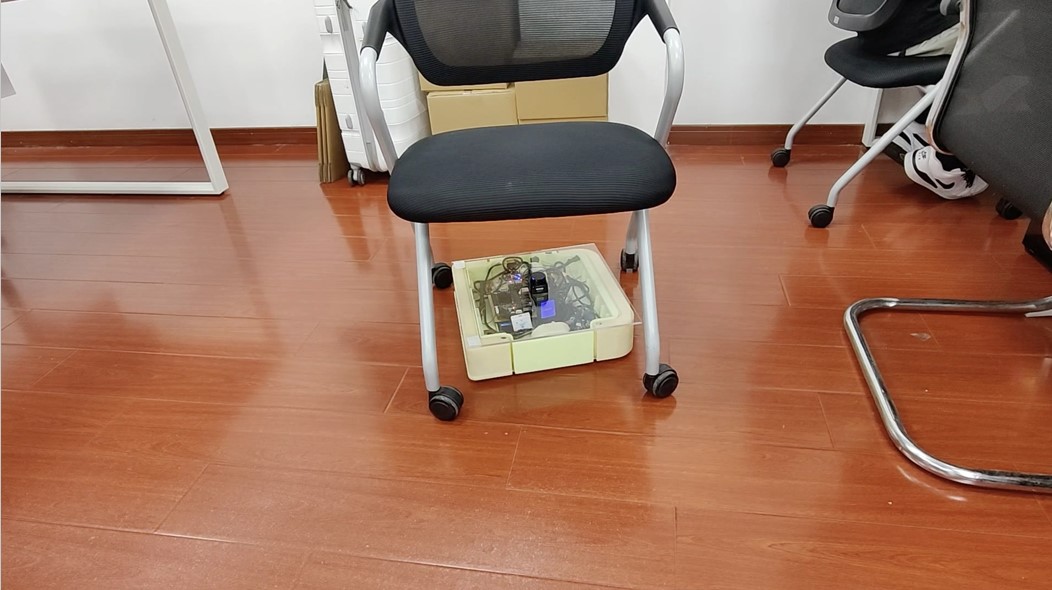}}
    \hfill
    \subfigure[Sparse obstacles]{\includegraphics[width=0.235\textwidth]{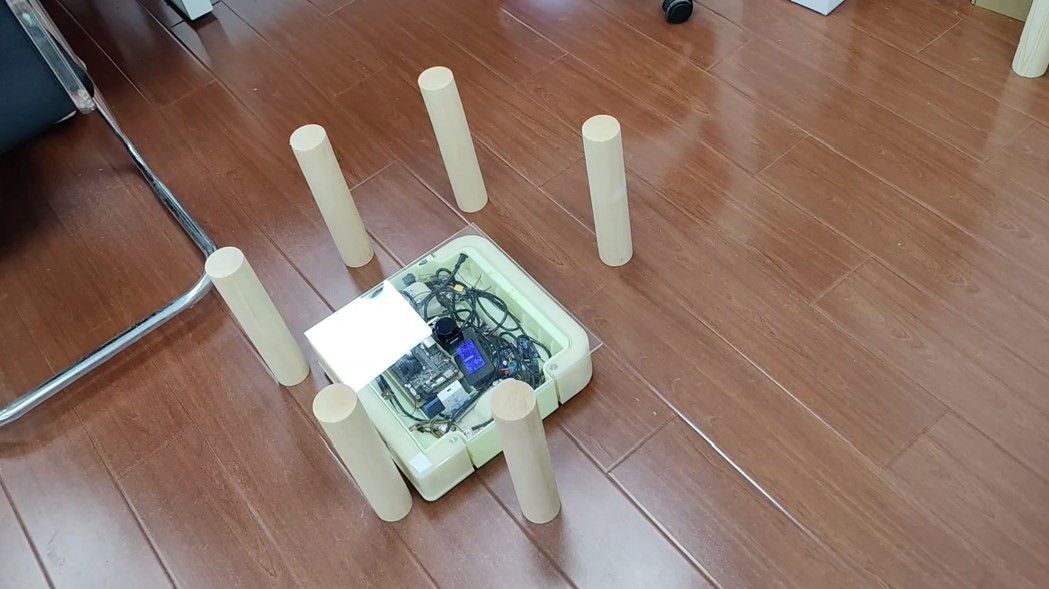}}
  \caption{The illustration of the real-world experiment scenarios. The figure shows the robot's initial position as it begins to escape from being stuck.}
  \label{fig:real-world}
  \vspace{-10pt}
\end{figure}

We utilized a self-developed real-world platform for testing.
The mobile platform is a square differential drive robot, measuring 35 cm by 35 cm.
It is equipped with a 2D Lidar for environmental perception and a Horizon X3 SDB computing platform for executing the algorithm.
Additionally, an IMU is integrated to provide rough localization information.
To validate the effectiveness of our algorithm in real-world scenarios, we conducted experiments in various escape scenarios, including narrow exit, long corridor, chair, and enclosures with sparse obstacles (as shown in Fig. \ref{fig:real-world}).
In addition to the listed scenarios, we also tested other scenarios and compared the planning results with ROS Navigation.
Notably, our algorithm can handle dynamic changes in the environment; due to our end-to-end architecture, it does not require re-planning and can respond to changes in real-time.
Also, these test scenarios were not encountered during training. The complete testing process can be viewed in the supplementary video.
The experimental results demonstrate that our algorithm successfully escaped in all scenarios, showcasing its robustness and generalization capability.

\section{Conclusion}
In this paper, we propose an end-to-end escape framework based on reinforcement learning. 
We employ a hybrid training strategy to accelerate the robot’s learning process. 
By utilizing real-time point cloud data, we implemented an efficient action mask that provides priors on collision-free action distributions.
Extensive comparative experiments demonstrate that our method consistently achieves higher success rates than traditional approaches and existing RL methods.
Real-world experiments also confirmed the effectiveness and generalization capability of our system.

Our algorithm focuses on escape tasks and uses an end-to-end approach that results in effective escape trajectories, but it compromises trajectory smoothness and length optimality. In the future, we aim to improve these aspects to achieve smoother and more optimal trajectories.
\clearpage


\end{document}